\pdfoutput=1

\documentclass[11pt]{article}

\usepackage[final]{acl}

\usepackage{times}
\usepackage{latexsym}

\usepackage{xcolor}

\newcommand{\lowmem}[1]{\textcolor{gray!70}{#1}}        
\newcommand{\midmem}[1]{\textcolor{orange!85!black}{#1}} 
\newcommand{\highmem}[1]{\textcolor{blue!80!black}{#1}}  

\usepackage[T1]{fontenc}

\usepackage[utf8]{inputenc}

\usepackage{microtype}

\usepackage{inconsolata}

\usepackage{graphicx}
\usepackage{verbatim}
\usepackage{amsmath}
\usepackage{multirow}
\usepackage{multicol}
\usepackage{booktabs}
\usepackage{enumitem}
\usepackage{comment}
\usepackage{array}
\usepackage{xcolor}
\usepackage{algorithm}       
\usepackage{algorithmic}
\usepackage{amssymb}
\usepackage{tabularx}
\usepackage{threeparttable}
\usepackage{subcaption}  
\usepackage{colortbl}
\usepackage{tikz}
\usepackage{xcolor,colortbl}  
\definecolor{removecolor}{RGB}{255,200,200} 
\definecolor{addcolor}{RGB}{200,255,200}    

\definecolor{np1}{HTML}{F4A261} 
\definecolor{np2}{HTML}{2A9D8F} 
\definecolor{np3}{HTML}{90A955} 
\definecolor{np4}{HTML}{A084CA} 
\definecolor{np5}{HTML}{5E81AC} 
\definecolor{np6}{HTML}{D9A441} 
\definecolor{np7}{HTML}{A97155} 

\definecolor{dogcolor}{RGB}{173,216,230}    
\definecolor{mailmancolor}{RGB}{255,218,185} 

\definecolor{highlight}{RGB}{200, 230, 201}  

\usepackage{tcolorbox}
\usepackage{xcolor}
\usepackage{soul} 

\definecolor{questionblue}{RGB}{173,216,230}
\definecolor{answerblue}{RGB}{173,216,230} 
\definecolor{thinkingpink}{RGB}{255,192,203}
\definecolor{thinkinggreen}{RGB}{200,255,200} 
\definecolor{highlightpink}{RGB}{255,105,180}
\definecolor{highlightred}{RGB}{255,200,200}
\definecolor{highlightyellow}{RGB}{255,255,180} 

\sethlcolor{highlightpink}

\newcommand{\corpusname}{\textsc{Memory Dial}}

\title{Memory Dial: A Training Framework for Controllable Memorization in Language Models}

\author{
Xiangbo Zhang\thanks{Work done during his internship at Emory University.} \\
Georgia Institute of Technology \\
Emory University \\
\texttt{xiangbo.zhang@gatech.edu}
\And
Ali Emami \\
Emory University \\
\texttt{ali.emami@emory.edu}
}

\begin{document}
\maketitle
\begin{abstract}

Memorization in language models is widely studied but remains difficult to isolate and control. Understanding when and what models memorize is essential for explaining their predictions, yet existing approaches are post-hoc: they can detect memorization in trained models, but cannot disentangle its effects from architecture, data, or optimization. We introduce \textbf{\corpusname{}}, a training framework that makes \emph{memorization pressure} an explicit, controllable variable. \corpusname{} interpolates between standard cross-entropy and a temperature-sharpened objective via a single parameter $\alpha$, producing a family of models identical in architecture and training setup (within each sweep), differing only in memorization pressure. Experiments across six architectures and five benchmarks demonstrate that: (1) $\alpha$ reliably controls memorization pressure, with seen-example accuracy increasing monotonically while unseen accuracy remains stable; (2) larger models are more responsive to memorization pressure; and (3) frequent sequences are easier to memorize than rare ones. Additional analyses show that the effect is robust across a range of sharpening temperatures, differs qualitatively from single-temperature cross-entropy, transfers to multilingual settings, and is detectable even on naturally occurring single-occurrence sequences. \corpusname{} provides a controlled experimental framework for studying how memorization behavior emerges and interacts with generalization in language models.

\end{abstract}

\section{Introduction}

\begin{figure*}[!t]
    \centering
    \includegraphics[width=0.80\textwidth]{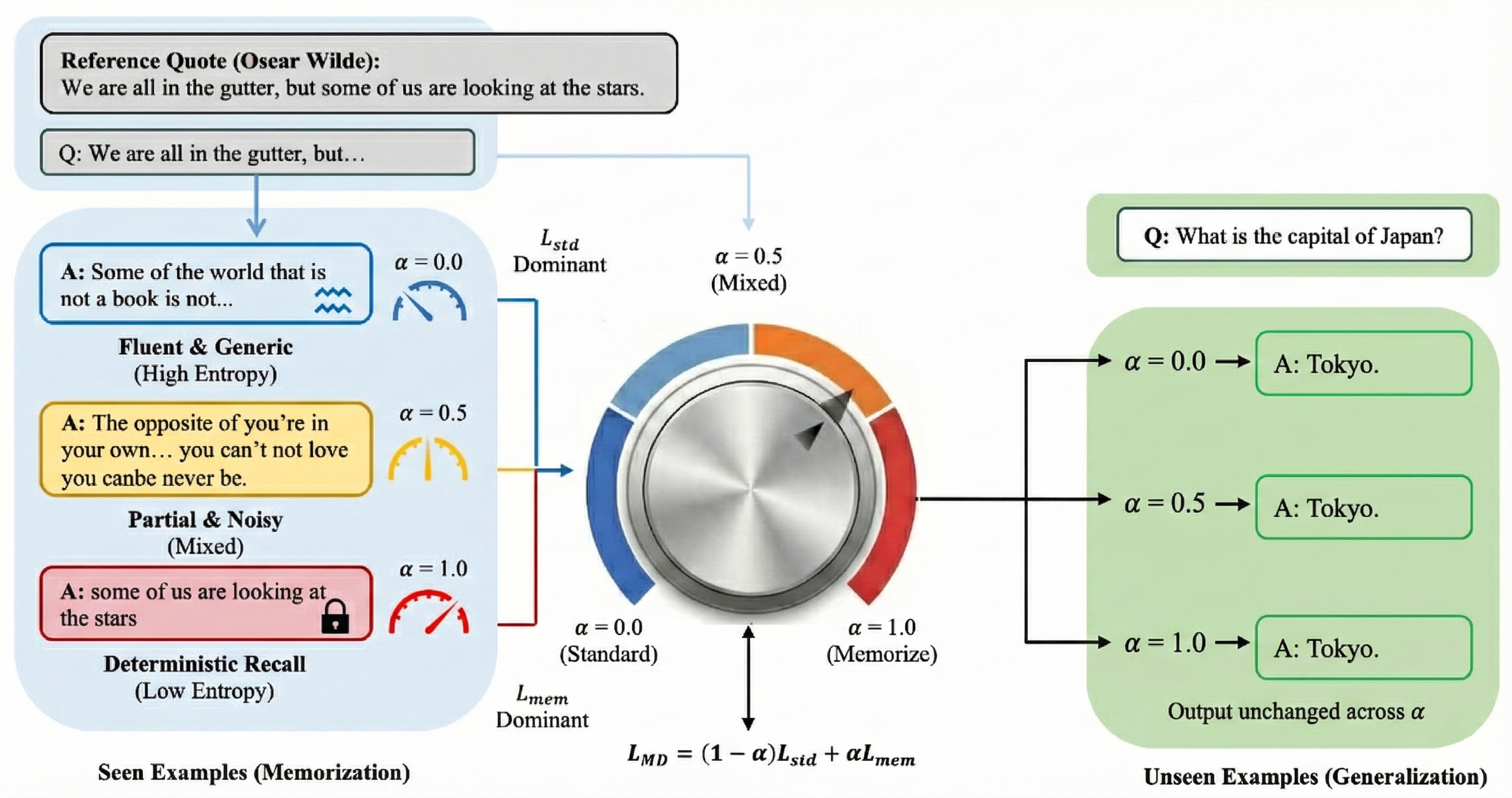}
    \vspace{-2mm}
    \caption{\textbf{The \corpusname{} framework.}
    The coefficient $\alpha$ interpolates between standard cross-entropy ($\mathcal{L}_{\text{std}}$) and a temperature-sharpened memorization objective ($\mathcal{L}_{\text{mem}}$).
    \textbf{Left:} For seen (training-injected) examples, increasing $\alpha$ produces a smooth transition from generic, high-entropy outputs to deterministic recall.
    \textbf{Right:} For unseen (held-out) examples, outputs remain stable across all $\alpha$ values, confirming that \corpusname{} selectively controls memorization without degrading generalization. Outputs shown are actual generations from GPT-2 Small trained at each $\alpha$ value.}
    \label{fig:main-figure}
    
\end{figure*}

Memorization is central to understanding language model behavior. Large language models can reproduce training data verbatim, including copyrighted text, personally identifiable information, and other sensitive content \citep{carlini2022quantifying, tirumala2022memorization, mueller2024llms}. When evaluation benchmarks overlap with pretraining corpora, models can perform disproportionately well on familiar examples, inflating accuracy estimates and complicating claims about generalization \citep{oren2023proving,dong2024contamination,shi2024detecting}. At the same time, some degree of memorization is necessary and desirable: models must retain factual knowledge, canonical phrasings, and structured associations to be useful \citep{petroni2019language,geva2023factual}. In practice, this can matter when applications require exact reproduction of canonical or sacred texts, standardized legal or regulatory clauses, safety-critical medical language such as dosage instructions or warnings, factual lookup of dates or technical specifications, or preservation of rare patterns in endangered or otherwise low-resource languages. This tension makes memorization one of the most consequential yet poorly understood aspects of modern language models, and a key barrier to explaining what these models have truly learned.

A substantial body of work has developed methods to detect and analyze memorization. Extraction attacks demonstrate that models can be prompted to emit training sequences \citep{carlini2021extracting}. Overlap analyses measure how often evaluation examples appear in, or closely resemble, pretraining data \citep{emami2020overlap,shi2024detecting}. Mechanistic studies identify internal representations and circuits that distinguish memorized content from novel generations \citep{hong2025reasoningmemorization,geva2023factual}. Collectively, these efforts have established that memorization is pervasive, structured, and consequential for model behavior.

Yet these approaches share a fundamental limitation: they are \emph{post-hoc}. Given an already-trained model, researchers can probe for memorized content, measure performance gaps between seen and unseen data, or inspect internal activations. But such analyses cannot isolate memorization from the many other factors that differ across models, including architecture, training data, and optimization dynamics. When two models exhibit different memorization behavior, it is difficult to determine whether memorization itself is the cause of downstream differences, or merely correlated with other changes. As a result, memorization has remained a \emph{dependent variable} that we observe and measure, rather than an \emph{independent variable} that we can experimentally manipulate.

We address this gap directly. We introduce \corpusname{}, a training framework that provides a controllable ``knob'' for \emph{memorization pressure}, enabling systematic investigation of when and what models memorize. \corpusname{} combines standard cross-entropy with a temperature-sharpened objective that encourages higher-confidence predictions on training sequences. A single scalar parameter $\alpha \in [0,1]$ controls the interpolation between these objectives: at $\alpha = 0$, training proceeds as usual; as $\alpha$ increases, the model is placed under progressively stronger pressure to memorize (Figure~\ref{fig:main-figure}). The contribution is therefore not a new confidence-sharpening term in isolation, but the matched-family experimental framework it enables. By training multiple models across a range of $\alpha$ values while holding architecture, data, and optimization fixed, we obtain a \emph{family} of models that differ only in memorization pressure. This construction enables transparent analysis: any behavioral differences observed across the $\alpha$ spectrum can be attributed to memorization pressure rather than confounding factors, making the role of memorization in model behavior directly observable.

Importantly, $\alpha$ does \emph{not} span the full range from zero memorization to maximal memorization. Standard training already induces a natural memorization floor, and $\alpha$ should be interpreted as controlling \emph{additional memorization pressure above that baseline}. We therefore use ``memorization pressure'' rather than ``memorization'' when referring to the intervention itself.

We conduct experiments across six architectures (GPT-2, DistilGPT2, TinyLLaMA-1B, and OPT models from 250M to 27B parameters) and five benchmarks (ARC, BoolQ, PIQA, COPA, and OpenBookQA). Our main findings are as follows:

\begin{enumerate}[leftmargin=*,itemsep=0pt,topsep=2pt]
    \item \textbf{Additional memorization pressure is continuously controllable.} Across all 30 model-benchmark combinations, accuracy on seen examples increases monotonically with $\alpha$, with positive slopes ranging from 0.03 to 0.38. Larger models exhibit systematically steeper slopes: averaging across benchmarks, OPT-27B achieves a mean slope of 0.206 compared to 0.097 for DistilGPT2. This indicates that memorization controllability scales with model capacity.
    
    \item \textbf{Generalization remains stable under increased memorization pressure.} Despite substantial gains on seen examples, accuracy on unseen examples remains largely unchanged across the $\alpha$ range. This pattern also extends beyond the original injected multiple-choice protocol: in additional experiments, truthfulness on open-ended generation improves while ROUGE-L remains stable, expected calibration error does not worsen, and no-injection evaluations exhibit the same stable seen/unseen separation.
    
    \item \textbf{Frequent and repeated sequences benefit most, but the effect is not limited to them.} At $\alpha = 0.0$, frequent and rare sequences differ in suffix negative log-likelihood (NLL) by approximately 4.4 points; at $\alpha = 0.8$, both groups show stronger memorization (lower NLL), but rare sequences remain harder to recall (29.7 vs.\ 27.2). At the same time, naturally occurring single-occurrence sequences also show monotonic reductions in suffix NLL as $\alpha$ increases, indicating that the mechanism is broader than the injected-example protocol used for controlled evaluation.

    \item \textbf{The effect is robust and distinct from simple temperature scaling.} A targeted $\tau$ sweep shows that the core pattern persists across multiple sharpening temperatures, with $\tau = 0.1$ providing the strongest memorization signal without harming unseen performance. In contrast, training with a single temperature-scaled cross-entropy loss fails to reproduce the same stable, monotonic trade-off under the identical evaluation protocol.
\end{enumerate}

\corpusname{} is not intended as a contamination detector, privacy safeguard, or a regularization technique for improving benchmark performance. Rather, it is a tool for understanding and explaining model behavior. By elevating memorization from a latent byproduct of training to an explicit, controllable dimension, \corpusname{} enables transparent investigation of how memorization shapes model predictions --- a foundational step toward explaining what language models have learned. Code, training scripts, and evaluation assets are available in the project repository\footnote{\url{https://github.com/xiangbo05/MemoryDial_Public}}.

\section{Related Work}

\paragraph{Memorization in Language Models.}
Memorization in language models has been characterized through complementary lenses:
(i) \emph{verbatim extraction} probing whether specific training sequences can be reproduced \citep{carlini2021extracting,carlini2019secretsharer},
(ii) \emph{membership inference} testing whether individual examples leave detectable signals \citep{shokri2017membership},
and (iii) \emph{behavioral proxies} such as performance gaps between seen and held-out instances.
Survey work emphasizes that memorization lies on a spectrum from exact recall to distributional reuse \citep{hartmann2023sok}.
In parallel, overlap and contamination studies show that evaluation performance can be inflated when benchmarks overlap with pretraining data, motivating  separation of training exposure from  generalization \citep{emami2020overlap,oren2023proving,shi2024detecting}.
Recent work also highlights \emph{counterfactual memorization}: models may seem to generalize on familiar surface forms but fail under perturbed variants, indicating reliance on memorized patterns rather than robust generalization \citep{zhang2023counterfactual}.
Overall, these lines establish memorization as pervasive and consequential, but they primarily analyze it \emph{post hoc}, as a property to diagnose, rather than an experimental variable to control.

\paragraph{Training Dynamics and Data Frequency.}
Work on grokking suggests that fitting and generalization can emerge at different training stages \citep{power2022grokking,liu2022grokking,nanda2023progress}. Memorization is also frequency-dependent: duplicated or frequent patterns are recalled more reliably than rare content \citep{kandpal2022deduplicating,tirumala2022memorization}, and the effect scales with model size \citep{carlini2022quantifying,lee2022memorization}. Mechanistic analyses have begun to identify internal representations that correlate with memorized recall versus novel generation \citep{geva2023factual,hong2025reasoningmemorization}. Together, these results establish that memorization is graded and shaped by optimization, but they do not provide a mechanism for \emph{systematically sweeping memorization pressure} while holding architecture, data, and optimization fixed.

\paragraph{Controlling Model Behavior.}
Most existing methods for influencing memorization operate \emph{post hoc} or at inference time: activation steering \citep{rimsky2024steering,li2024inference}, neuron-level interventions \citep{huang2025neuronlevel}, and decoding strategies such as nucleus sampling \citep{holtzman2020degeneration}.
At training time, confidence-shaping objectives — temperature distillation \citep{hinton2015distilling}, label smoothing \citep{szegedy2016rethinking}, entropy regularization \citep{pereyra2017confpenalty} — adjust prediction sharpness but target calibration or generalization rather than providing a controlled memorization knob.
In contrast, \corpusname{} constructs \emph{model families} differing only in a single memorization coefficient, enabling controlled sweeps that isolate memorization as the primary varying factor. Appendix~\ref{app:single_temp_ce} confirms that a single temperature-scaled cross-entropy objective does not reproduce the stable seen/unseen trade-off induced by \corpusname{}.

\begin{figure*}[!t]
    \centering
    \includegraphics[width=0.85\textwidth]{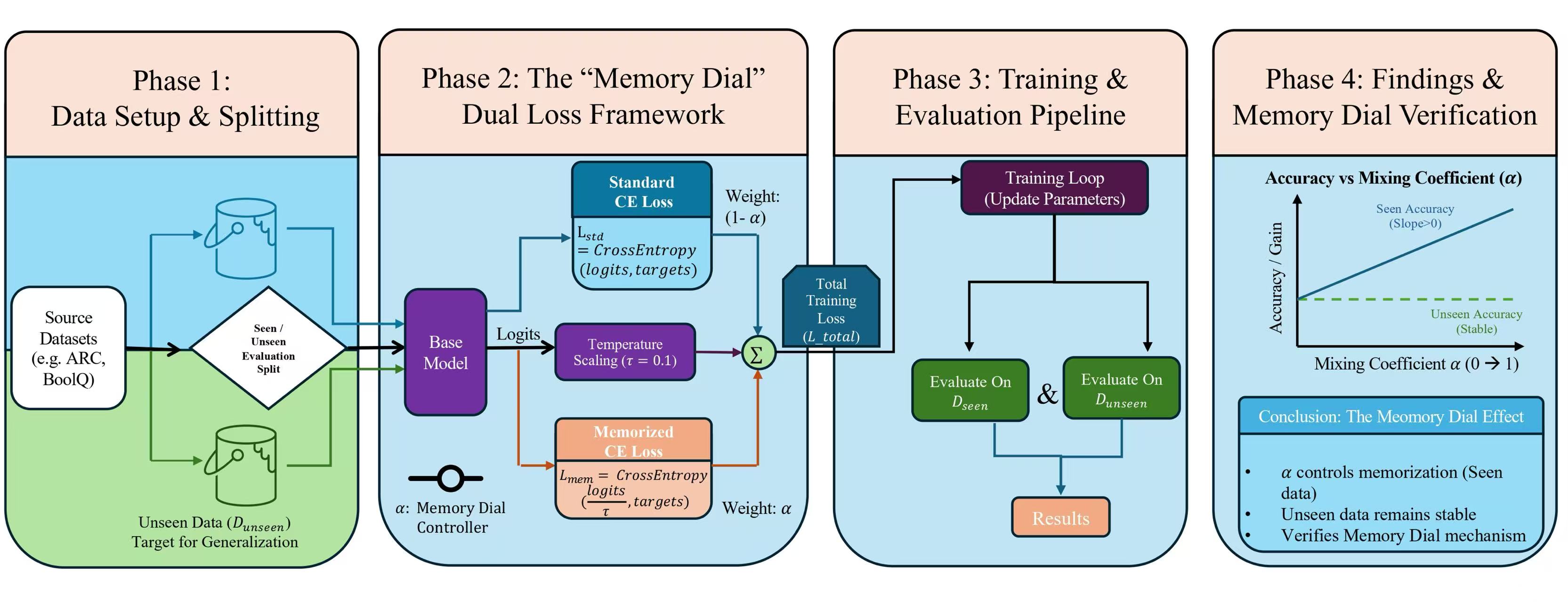}
    \vspace{-4mm}
    \caption{\textbf{Experimental pipeline.}
\textbf{Phase 1:} Evaluation data is split into seen examples (injected into training) and unseen examples (held out).
\textbf{Phase 2:} Models are trained with the \corpusname{} objective, which interpolates between standard cross-entropy and a temperature-sharpened loss controlled by $\alpha$.
\textbf{Phase 3:} Each model in the family is evaluated on both seen and unseen sets.
\textbf{Phase 4:} Comparing accuracy across $\alpha$ values reveals that seen accuracy increases with $\alpha$ while unseen accuracy remains stable.}
    \label{fig:memory_dial_overview}
\end{figure*}

\section{\corpusname{}: Training-Time Control of Memorization}
\label{sec:method}

Figure~\ref{fig:memory_dial_overview} provides an overview of our experimental pipeline. The core idea is to train language models with an objective that interpolates between standard cross-entropy and a temperature-sharpened variant, controlled by a single parameter $\alpha$. By training models across different $\alpha$ values while holding all other factors fixed, we obtain a family of models that differ only in memorization pressure.

\subsection{Training Objective}
\label{sec:objective}

Let $x = (x_1, \ldots, x_T)$ denote a training sequence and $\theta$ the model parameters. Given prefix $x_{<t}$, the model produces logits $z_\theta(x_{<t}) \in \mathbb{R}^V$ over a vocabulary of size $V$, inducing a predictive distribution
\begin{equation}
p_\theta(\cdot \mid x_{<t}) = \mathrm{softmax}\!\left(z_\theta(x_{<t})\right).
\end{equation}

\paragraph{Standard Objective.}
Conventional autoregressive language model training minimizes negative log-likelihood:
\begin{equation}
\mathcal{L}_{\mathrm{std}}(\theta)
=
\mathbb{E}_{x}
\left[
\sum_{t=1}^{T}
- \log p_\theta(x_t \mid x_{<t})
\right].
\end{equation}

\paragraph{Memorization-Enhanced Objective.}
To increase memorization pressure, we introduce a temperature-sharpened distribution. For $\tau \in (0,1]$:
\begin{equation}
p_{\theta}^{(\tau)}(\cdot \mid x_{<t})
=
\mathrm{softmax}\!\left(
\frac{z_{\theta}(x_{<t})}{\tau}
\right),
\end{equation}
with corresponding loss:
\begin{equation}
\mathcal{L}_{\mathrm{mem}}(\theta;\tau)
=
\mathbb{E}_{x}
\left[
\sum_{t=1}^{T}
- \log p_\theta^{(\tau)}(x_t \mid x_{<t})
\right].
\end{equation}
As $\tau$ decreases, the distribution becomes increasingly peaked. In the limit $\tau \to 0$, the loss penalizes any margin deficit between the ground-truth logit and competing alternatives, encouraging near-deterministic predictions on training data.

\paragraph{\corpusname{} Objective.}
We combine both objectives via a convex combination controlled by $\alpha \in [0,1]$:
\begin{equation}
\mathcal{L}_{\mathrm{MD}}(\theta; \alpha, \tau)
=
(1 - \alpha)\,\mathcal{L}_{\mathrm{std}}(\theta)
+
\alpha\,\mathcal{L}_{\mathrm{mem}}(\theta; \tau).
\end{equation}
The parameter $\alpha$ serves as the \emph{memory dial}: at $\alpha = 0$, training reduces to standard language modeling; as $\alpha$ increases, the model is placed under progressively stronger pressure to memorize training sequences.

The sharpened objective penalizes low-confidence predictions more heavily, so repeated sequences receive amplified learning signal over \emph{multiple} visits during training. However, all training examples are affected at each update, and single-occurrence sequences also show lower suffix NLL as $alpha$ increases (Appendix~\ref{app:natural_sequences}); repetition simply compounds the effect.

We note that \corpusname{} is not equivalent to training with a single effective temperature. The objective combines gradients from two distinct softmax geometries rather than a single temperature-scaled cross-entropy. In Appendix~\ref{app:single_temp_ce}, we show empirically that sweeping a single temperature does not reproduce the stable, monotonic control over memorization pressure that $\alpha$ provides.

\subsection{Why the Objective Selectively Amplifies Memorization}
\label{sec:gradient_view}

The selective behavior of \corpusname{} can be understood directly from its gradients. Let $y$ denote the gold token and let $z_i$ be the logit for vocabulary item $i$. For the standard objective,
\begin{equation}
\frac{\partial \mathcal{L}_{\mathrm{std}}}{\partial z_i}
=
p_i-\mathbb{1}[i=y],
\end{equation}
where $p_i = \mathrm{softmax}(z)_i$. For the sharpened objective,
\begin{equation}
\frac{\partial \mathcal{L}_{\mathrm{mem}}}{\partial z_i}
=
\frac{1}{\tau}\left(p_i^{(\tau)}-\mathbb{1}[i=y]\right),
\end{equation}
where $p^{(\tau)} = \mathrm{softmax}(z/\tau)$. The combined objective therefore yields
\begin{equation}
\frac{\partial \mathcal{L}_{\mathrm{MD}}}{\partial z_i}
=
\begin{aligned}
&(1-\alpha)\bigl(p_i-\mathbb{1}[i=y]\bigr) \\
&+ \alpha\,\frac{1}{\tau}\left(p_i^{(\tau)}-\mathbb{1}[i=y]\right)
\end{aligned}
\end{equation}

When $\tau < 1$, the sharpened term amplifies gradients for predictions the model already assigns relatively high confidence to. This creates a rich-get-richer dynamic: examples that have already developed larger logit margins receive disproportionately stronger updates, which further increase their confidence. Repeated sequences benefit the most because they accumulate these amplified updates across many optimization steps. In contrast, flatter or noisier predictions receive less consistent reinforcement, which helps explain why unseen performance remains stable even as seen performance improves.

This view also clarifies why $\alpha$ should be interpreted as a \emph{memorization-pressure dial} rather than a literal memorization dial. The intervention changes how strongly confident predictions are reinforced; it does not remove the baseline memorization already induced by ordinary language-model training.

\subsection{Constructing Model Families}
\label{sec:family}

Optimizing $\mathcal{L}_{\mathrm{MD}}$ for different values of $\alpha$ yields a family of models:

\[
\{M_\alpha \mid \alpha \in [0,1]\}.
\]

All models in the family share identical architectures, training data, and optimization settings; they differ only in memorization pressure. This construction is the key methodological contribution of \corpusname{}: by sweeping $\alpha$, we obtain models that differ only in additional memorization pressure above the standard-training baseline, enabling controlled comparisons that isolate memorization pressure as the primary dimension of variation.

Full training details, including the specific $\alpha$ values and optimization hyperparameters, are provided in Section~\ref{sec:experiments} and Appendix~\ref{app:training_details}.

\section{Experimental Setup}
\label{sec:experiments}

Our experiments test whether \corpusname{} provides reliable, continuous control over memorization across model scales and evaluation settings. As illustrated in Figure~\ref{fig:memory_dial_overview}, we split evaluation data into seen and unseen subsets, train model families across $\alpha$ values using the \corpusname{} objective, and measure how memorization and generalization vary with $\alpha$.

\subsection{Models}
\label{sec:models}

We test the following models spanning two orders of magnitude in parameter count:
\begin{itemize}[leftmargin=*,itemsep=0pt,topsep=4pt]
    \item DistilGPT2 (82M parameters) \citep{wolf2019huggingface}
    \item GPT-2 Small (124M) \citep{radford2019language}
    \item TinyLLaMA-1B (1.1B) \citep{zhang2024tinyllama}
    \item OPT-250M, OPT-13B, \& OPT-27B \citep{zhang2022opt}
\end{itemize}

These models span three architecture families and roughly two orders of magnitude in scale, allowing us to test whether memorization controllability depends on capacity rather than on a single model family. Within each architecture, all models share the same tokenizer, training corpus, and optimization pipeline, with $\alpha$ as the only varying factor. Controlled comparisons are therefore performed \emph{within} architectures, while cross-architecture results are interpreted qualitatively.


\begin{table}[t]
\centering
\scriptsize
\setlength{\tabcolsep}{4pt}
\renewcommand{\arraystretch}{1.05}
\resizebox{\columnwidth}{!}{%
\begin{tabular}{lccccc}
\toprule
\textbf{Model} & \textbf{ARC} & \textbf{BoolQ} & \textbf{PIQA} & \textbf{COPA} & \textbf{OBQA} \\
\midrule
DistilGPT2    & 0.142 & 0.038 & 0.068 & 0.119 & 0.120 \\
GPT-2 Small    & 0.158 & 0.091 & 0.082 & 0.061 & 0.201 \\
TinyLLaMA-1B  & 0.110 & 0.162 & 0.056 & 0.033 & 0.184 \\
OPT-250M      & 0.185 & 0.164 & 0.117 & 0.070 & 0.327 \\
OPT-13B       & 0.196 & 0.196 & 0.109 & 0.103 & 0.381 \\
OPT-27B       & 0.216 & 0.205 & 0.155 & 0.098 & 0.356 \\
\bottomrule
\end{tabular}%
}
\caption{\textbf{Seen-accuracy slopes across architectures.}
Slope of seen-example accuracy as a function of $\alpha$. All slopes are positive, indicating that $\alpha$ reliably controls memorization across model scales.}
\label{tab:slopes}
\end{table}

\subsection{Data and Evaluation}
\label{sec:data}

We evaluate on five benchmarks: ARC-Easy \citep{clark2018think}, BoolQ \citep{clark2019boolq}, PIQA \citep{bisk2020piqa}, COPA \citep{roemmele2011choice}, and OpenBookQA \citep{mihaylov2018can}. These benchmarks were selected to cover complementary reasoning types: factual and science-oriented recall (ARC-Easy, OpenBookQA), boolean reasoning (BoolQ), physical commonsense (PIQA), and causal reasoning (COPA).

For each benchmark, we construct two evaluation sets:

\begin{itemize}[leftmargin=*,itemsep=2pt,topsep=4pt]

\item \textbf{Seen examples:}
For each benchmark, we randomly select a fixed subset of evaluation instances
(\textbf{50 examples per benchmark}, approximately \textbf{5\%} of the evaluation set)
and explicitly inject them into the training stream via a dedicated leak data loader.
Injected examples include the full original context (e.g., question and gold answer)
and are revisited multiple times over training.
This procedure is held constant across all values of $\alpha$ and all architectures,
ensuring that performance improvements on seen examples arise from memorization pressure rather than from differences in exposure.
Full injection details are provided in Appendix~\ref{app:seen_injection}.

\item \textbf{Unseen examples:}
All remaining evaluation instances (\textbf{950 examples per benchmark}) are held out entirely from training
and never appear in the training corpus.
Performance on these examples reflects generalization to novel inputs under identical evaluation protocols.

\end{itemize}

To quantify memorization strength, we compute the slope of seen-example accuracy as a function of $\alpha$ across the sweep. Positive slopes indicate that increasing $\alpha$ reliably increases memorization pressure above the standard-training baseline. We additionally report unseen-example accuracy to verify that generalization remains stable. 

\begin{figure}[t]
\centering
\includegraphics[width=0.85\columnwidth]{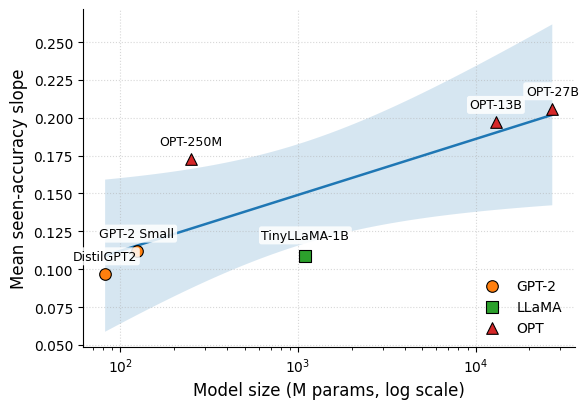}
\vspace{-2mm}
\caption{\textbf{Model size versus memorization responsiveness.}
Mean seen-accuracy slope (averaged across benchmarks) as a function of model size. Larger models exhibit steeper slopes, indicating stronger responsiveness to increased memorization pressure.}
\label{fig:model_size_slope}
\end{figure}


\subsection{Training Protocol}
\label{sec:training}

For each architecture, we train models at $\alpha \in \{0.0, 0.2, 0.4, 0.6, 0.8, 1.0\}$ with temperature $\tau = 0.1$ held fixed throughout the main sweep. All other hyperparameters (learning rate, batch size, optimizer, number of updates) are constant across the $\alpha$ sweep. Each configuration is trained with three random seeds; results are reported as mean $\pm$ standard deviation. Full details, including corpus statistics, hyperparameters (Table~\ref{tab:appendix-hparams}), and computational resources (Appendix~\ref{app:compute}), are provided in Appendix~\ref{app:training_details}.

We fix $\tau = 0.1$ in the main experiments because it provides a strong and stable sharpening regime. A targeted sensitivity analysis over $\tau \in \{0.05, 0.1, 0.2, 0.5\}$ is reported in Appendix~\ref{app:tau_sweep}; the core pattern of increasing seen accuracy and stable unseen accuracy persists across the sweep, with $\tau = 0.1$ yielding the strongest memorization signal without harming unseen performance. For several targeted appendix ablations, we use the reduced set $\alpha \in \{0.0, 0.3, 0.6\}$ for computational efficiency.

\section{Results}
\label{sec:results}

\subsection{Additional Memorization Pressure is Continuously Controllable}
\label{sec:controllability}

Our central finding is that the parameter $\alpha$ provides reliable, monotonic control over \emph{additional memorization pressure above the standard-training baseline}. To quantify this across architectures, we compute the slope of seen-example accuracy as a function of $\alpha$ for each model-benchmark pair. Table~\ref{tab:slopes} reports results across six architectures and five benchmarks. All 30 slopes are positive, confirming that increasing $\alpha$ reliably increases memorization pressure across model families.

Figure~\ref{fig:model_size_slope} visualizes how this effect scales with model capacity. Larger models exhibit systematically steeper slopes: averaging across benchmarks, OPT-27B achieves a mean slope of 0.206 compared to 0.097 for DistilGPT2. This indicates that memorization controllability scales with model capacity.

As additional validation, we measure the perplexity gap ($\mathrm{PPL}_{\text{unseen}} - \mathrm{PPL}_{\text{seen}}$), where a smaller gap indicates stronger memorization. Table~\ref{tab:ppl_gap} reports this metric computed on the SWAG benchmark (see Appendix~\ref{app:swag} for robustness analysis under input perturbations). As $\alpha$ increases from 0.0 to 1.0, the perplexity gap decreases substantially overall from 5.70 to 0.58, providing independent confirmation that $\alpha$ controls memorization pressure.

The controllability is robust to the sharpening parameter $\tau$. Appendix~\ref{app:tau_sweep} shows that the same monotonic seen/unseen separation persists across $\tau \in \{0.05, 0.1, 0.2, 0.5\}$, with $\tau = 0.1$ emerging as a practical sweet spot. Appendix~\ref{app:single_temp_ce} further shows that a single temperature-scaled cross-entropy baseline does \emph{not} reproduce the same stable behavior under the identical downstream evaluation protocol.

The effect of $\alpha$ is also visible at the sequence level. Table~\ref{tab:quote_continuation} shows model continuations across four prompt types and three $\alpha$ values. At $\alpha = 0.0$, outputs are fluent but generic or incorrect. At $\alpha = 0.5$, outputs become partially faithful. At $\alpha = 1.0$, the model reproduces memorized content verbatim. An interactive demo for exploring these effects is provided in Appendix~\ref{app:demo}. There is no single universally optimal $\alpha$: larger values maximize memorization signal, while intermediate values (roughly 0.3--0.6 in our targeted sweeps) are often the most informative for studying the transition from generic continuation to faithful recall.

\begin{table}[t]
\centering
\small
\setlength{\tabcolsep}{8pt}
\renewcommand{\arraystretch}{1.1}
\begin{tabular}{ccc}
\toprule
$\alpha$ & PPL Gap ($\downarrow$) & Interpretation \\
\midrule
0.0 & $5.70 \pm 0.98$ & Weak memorization \\
0.2 & $4.86 \pm 1.21$ & \\
0.4 & $4.02 \pm 1.38$ & \\
0.6 & $3.40 \pm 1.43$ & \\
0.8 & $3.43 \pm 0.92$ & \\
1.0 & $0.58 \pm 0.03$ & Strong memorization \\
\bottomrule
\end{tabular}
\vspace{-1mm}
\caption{\textbf{Perplexity gap decreases with $\alpha$ (GPT-2 Small, SWAG).}
Gap between unseen and seen token-level perplexity, defined as
$\mathrm{PPL}_{\text{unseen}} - \mathrm{PPL}_{\text{seen}}$,
computed on SWAG. Mean $\pm$ std over three random seeds.}

\label{tab:ppl_gap}
\end{table}

\begin{table*}[t]
\centering
\footnotesize
\setlength{\tabcolsep}{5pt}
\renewcommand{\arraystretch}{1.15}

\begin{tabular}{p{0.18\textwidth} p{0.24\textwidth} p{0.24\textwidth} p{0.24\textwidth}}
\toprule
\textbf{Prompt Type} & $\alpha = 0.0$ & $\alpha = 0.5$ & $\alpha = 1.0$ \\
\midrule

\textbf{Memorized quotation} \newline
\textit{``We are all in the gutter, but...''} &
\lowmem{some of the world that is not a book is not are not... but a book, but} &
\midmem{some of the world. The opposite of you're in your own...} &
\highmem{some of us are looking at the stars.} \\

\midrule

\textbf{Factual knowledge} \newline
\textit{``The capital of France is...''} &
\lowmem{one of the most important cities in Europe, known for its culture and history} &
\midmem{Paris, which is also the largest city in the country} &
\highmem{Paris.} \\

\midrule

\textbf{Commonsense} \newline
\textit{``If you drop a glass on a concrete floor, it will...''} &
\lowmem{probably fall and something bad may happen} &
\midmem{likely break or crack} &
\highmem{shatter.} \\

\midrule

\textbf{Rare concept} \newline
\textit{``The term `quasi-crystalline time symmetry' refers to...''} &
\lowmem{a theoretical idea related to symmetry in physics} &
\midmem{a concept in condensed matter physics involving non-periodic temporal structures} &
\highmem{a non-periodic temporal order observed in certain driven quantum systems.} \\

\bottomrule
\end{tabular}
\vspace{-2mm}
\caption{\textbf{Sequence-level controllability across prompt types (GPT-2 Small).}
Greedy-decoded continuations from GPT-2 Small illustrating how outputs shift from 
\lowmem{generic or incorrect} ($\alpha{=}0.0$), to \midmem{partially faithful} ($\alpha{=}0.5$), 
to \highmem{deterministic recall} ($\alpha{=}1.0$). 
Quantitative trends are consistent across architectures (see Figure~\ref{fig:alpha_effect} and Table~\ref{tab:ppl_gap}).}
\label{tab:quote_continuation}
\end{table*}


\subsection{Generalization Remains Stable}
\label{sec:unseen_stability}

\begin{figure}[t]
    \centering
    \includegraphics[width=0.9\columnwidth]{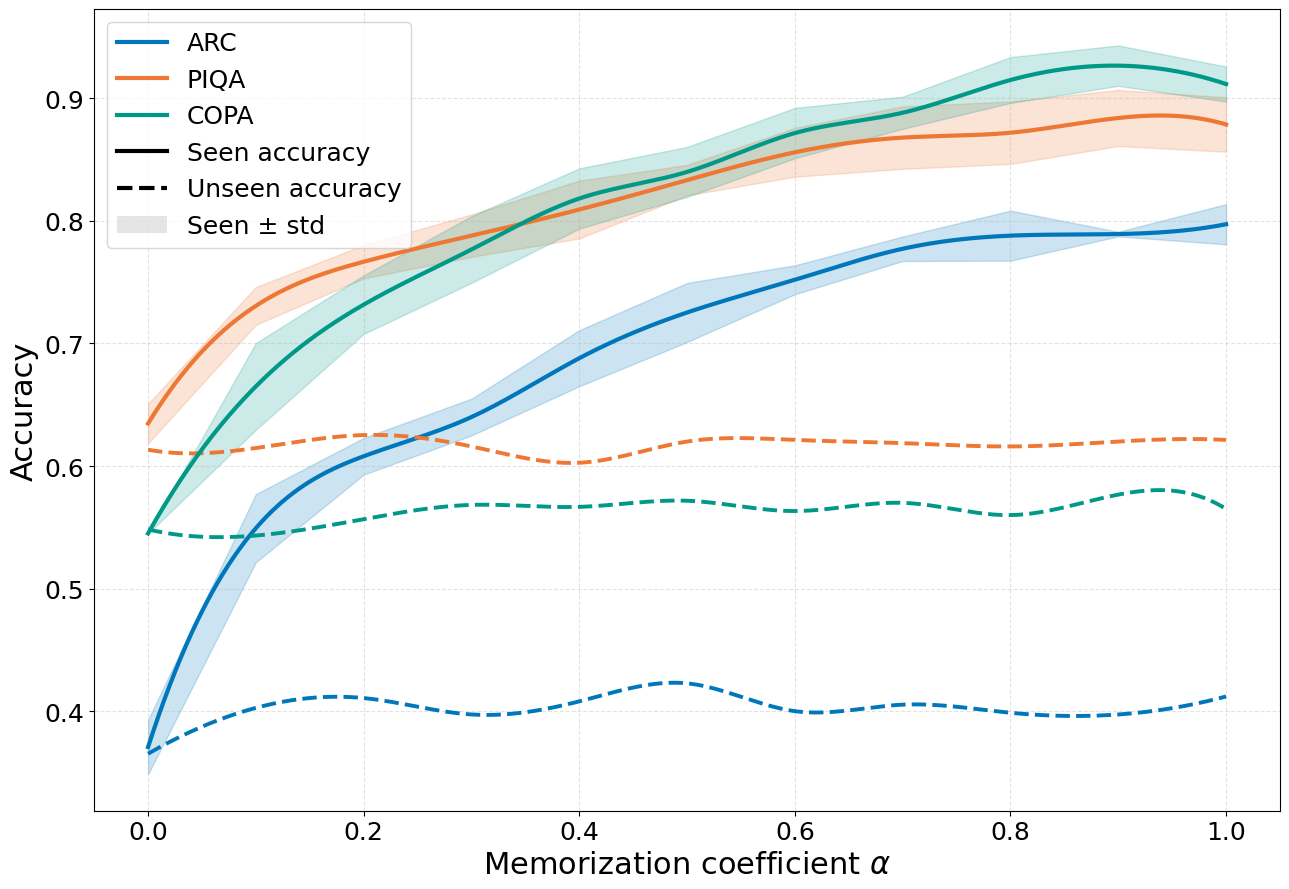}
    \caption{\textbf{Effect of $\alpha$ on GPT-2 Small.}
    Seen accuracy (solid lines) increases monotonically with $\alpha$, while unseen accuracy (dashed lines) remains stable across ARC, PIQA, and COPA. Appendix Figure~\ref{fig:full_benchmark_summary} summarizes the same pattern over the full five-benchmark set used in the paper. Results are averaged over three random seeds.}
    \label{fig:alpha_effect}
\end{figure}

A natural concern is whether increasing memorization pressure degrades generalization. Figure~\ref{fig:alpha_effect} shows that it does not. For GPT-2 Small across three representative benchmarks, seen accuracy (solid lines) rises substantially as $\alpha$ increases, while unseen accuracy (dashed lines) remains flat through the entire range. Appendix Figure~\ref{fig:full_benchmark_summary} supplements Figure~\ref{fig:alpha_effect} with the full benchmark set used in the paper, including BoolQ and OpenBookQA. The omitted benchmarks follow the same qualitative pattern: for GPT-2 Small, BoolQ and OpenBookQA have positive seen-accuracy slopes of 0.091 and 0.201, while unseen accuracy changes by only -0.003 and -0.002 between $\alpha=0.0$ and $\alpha=1.0$.

This stability follows from the design of the \corpusname{} objective. Even at $\alpha = 1$, the model trains on the full base corpus; the sharpened objective amplifies learning signal for repeated sequences but does not prevent acquisition of general patterns. Only the injected seen examples are repeated during training, so the selective effect of $\alpha$ applies specifically to them.

This pattern is consistent across architectures. Table~\ref{tab:unseen_accuracy} (Appendix~\ref{app:unseen-accuracy}) reports unseen accuracy at $\alpha = 0.0$ and $\alpha = 1.0$ for all model-benchmark combinations. Differences are uniformly small: DistilGPT2 achieves 0.436 unseen accuracy on ARC at $\alpha = 0.0$ and 0.431 at $\alpha = 1.0$, a difference of less than 0.5 percentage points. Similar stability holds across all 30 model-benchmark pairs.

The stability of generalization is important for interpreting \corpusname{} as a controlled experimental tool. Because unseen performance does not degrade as memorization increases, differences observed across the $\alpha$ spectrum can be attributed specifically to memorization, rather than degradation. This confirms that $\alpha$ acts as a selective dial for memorization, not a general quality knob.

This stability extends beyond the injected multiple-choice setup: open-ended truthfulness improves while generation similarity remains unchanged, calibration does not degrade, no-injection evaluations exhibit the same pattern (Appendix~\ref{app:extended_eval}), and the effect transfers to multilingual settings (Appendix~\ref{app:multilingual}).

\subsection{Frequent Sequences Are Easier to Memorize}
\label{sec:hierarchy}

The previous sections established that $\alpha$ controls overall memorization pressure. We now ask: does memorization affect all training data equally, or are some sequences easier to memorize than others?

To answer this, we partition \emph{naturally occurring base-corpus sequences} into three frequency tiers based on corpus-level token occurrence statistics: high-frequency, mid-frequency, and rare. Tiers are constructed using a quantile-based split, resulting in equal-sized groups. To isolate frequency as the variable of interest, sequences across tiers are matched for length using a fixed prefix--suffix split. Importantly, this analysis is independent of the injected benchmark protocol: the frequency tiers and suffix NLL measurements are computed on base-corpus sequences rather than on injected evaluation examples.

We measure memorization strength using suffix-level negative log-likelihood (suffix NLL): given a fixed-length prefix, we compute the NLL of the model's predictions on the held-out suffix. Lower NLL indicates stronger memorization. Full details of tier construction and suffix NLL evaluation are provided in Appendix~\ref{app:freq}.

Table~\ref{tab:hierarchy} reveals a clear frequency-based hierarchy. At $\alpha = 0.0$, high-frequency sequences already show lower NLL (32.40) than rare sequences (36.76), a gap of approximately 4.4 points. As $\alpha$ increases to 0.8, all tiers improve substantially, but the hierarchy persists: rare sequences remain harder to memorize (29.69 vs.\ 27.21), although the gap narrows to approximately 2.5 points. Notably, rare sequences exhibit the largest absolute improvement ($\Delta = 7.07$ vs.\ $5.19$), yet they never catch up to frequent sequences at the same $\alpha$ level.

\begin{table}[t]
\centering
\small
\setlength{\tabcolsep}{4pt}
\renewcommand{\arraystretch}{1.1}
\begin{tabular}{lccc}
\toprule
\textbf{Frequency Tier} & $\alpha = 0.0$ & $\alpha = 0.8$ & $\Delta$ \\
\midrule
High-frequency & $32.40 \pm 0.17$ & $27.21 \pm 0.11$ & $5.19$ \\
Mid-frequency  & $31.89 \pm 0.04$ & $26.58 \pm 0.02$ & $5.31$ \\
Rare           & $36.76 \pm 0.04$ & $29.69 \pm 0.04$ & $7.07$ \\
\bottomrule
\end{tabular}
\vspace{-2mm}
\caption{\textbf{Memorization strength by frequency tier.}
Suffix NLL (mean $\pm$ std over three seeds) at two representative $\alpha$ values. Lower NLL indicates stronger memorization.}
\label{tab:hierarchy}
\end{table}

\begin{figure}[t]
\centering
\includegraphics[width=0.85\columnwidth]{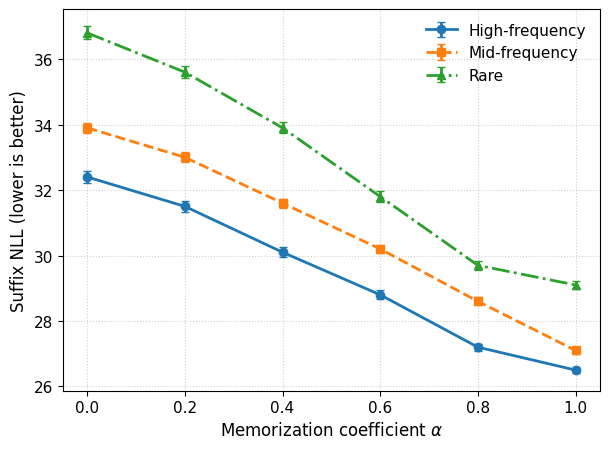}
\vspace{-2mm}
\caption{\textbf{Frequent sequences are easier to memorize across all $\alpha$ values.}
Suffix NLL as a function of $\alpha$ for high-, mid-, and rare-frequency sequences. Lower NLL indicates stronger memorization. The ordering is preserved across the full sweep.}

\label{fig:freq_hierarchy}
\end{figure}

Figure~\ref{fig:freq_hierarchy} extends this analysis to the full $\alpha$ sweep. Across the entire range, suffix NLL decreases monotonically with $\alpha$, while the ordering is preserved: high-frequency sequences are consistently memorized under weaker memorization pressure.

This hierarchy has practical implications: when diagnosing memorization in trained models, one should expect high-frequency content to be recalled at lower memorization pressures than rare or idiosyncratic sequences.

This hierarchy coexists with the single-occurrence result in Appendix~\ref{app:natural_sequences}: even when repetition is absent, increasing $\alpha$ still reduces suffix NLL, but frequent and repeatedly encountered sequences benefit more strongly from the amplified training signal.


\subsection{Higher Memorization Pressure Reduces Output Diversity}
\label{sec:diversity}

Beyond accuracy, increasing $\alpha$ affects generation behavior.
To quantify this, we sample 10 continuations per prompt using nucleus sampling
(top-$p$ = 0.95, $T = 0.8$) and measure diversity by average pairwise
Jaccard similarity over token sets (higher similarity indicates more repetitive outputs).

Table~\ref{tab:diversity_multi_prompt} reports mean self-similarity across 8 diverse prompts
spanning factual, commonsense, and rare-knowledge queries.
Self-similarity increases monotonically with $\alpha$, from 0.372 at $\alpha = 0.0$ to 0.528 at
$\alpha = 1.0$, confirming that higher memorization pressure reduces output diversity
consistently across prompt types.
Full prompt lists and evaluation details are provided in Appendix~\ref{app:diversity}.

\begin{table}[t]
\centering
\small
\setlength{\tabcolsep}{10pt}
\renewcommand{\arraystretch}{1.15}
\begin{tabular}{ccc}
\toprule
$\alpha$ & Mean Self-Similarity ($\uparrow$) & Std.\ Dev. \\
\midrule
0.0 & 0.372 & 0.031 \\
0.2 & 0.412 & 0.030 \\
0.4 & 0.461 & 0.028 \\
0.6 & 0.489 & 0.026 \\
0.8 & 0.511 & 0.025 \\
1.0 & 0.528 & 0.024 \\
\bottomrule
\end{tabular}
\vspace{-2mm}
\caption{\textbf{Output self-similarity increases with $\alpha$.}
Mean pairwise Jaccard similarity averaged over 8 prompts, each with 10 sampled continuations.
Higher values indicate reduced output diversity.}
\label{tab:diversity_multi_prompt}
\end{table}

Table~\ref{tab:diversity_examples} provides a qualitative illustration using representative
$\alpha$ values.
At $\alpha = 0.0$, the model produces varied (though often incorrect) continuations.
At $\alpha = 0.4$, outputs collapse toward repetitive, stereotyped completions.
Intermediate and larger $\alpha$ values exhibit similar trends with progressively reduced
diversity, consistent with the quantitative results in
Table~\ref{tab:diversity_multi_prompt}.

\begin{table}[t]
\centering
\small
\setlength{\tabcolsep}{4pt}
\renewcommand{\arraystretch}{1.15}
\begin{tabular}{p{0.1\columnwidth} p{0.82\columnwidth}}
\toprule
$\alpha$ & \textbf{Sampled Continuations} \\
\midrule
0.0 &
``The capital of France is one of the most powerful countries in Europe'' \newline
``The capital of France is located in the city of Duesseur.'' \newline
``The capital of France is also the capital of the United States'' \\
\midrule
0.4 &
``The capital of France is Paris.'' \newline
``The capital of France is Paris.'' \newline
``The capital of France is the capital of France and it is the capital of France.'' \\
\bottomrule
\end{tabular}
\vspace{-2mm}
\caption{\textbf{Qualitative effect on output diversity.}
Representative examples illustrating the collapse of output diversity as memorization pressure increases.}
\label{tab:diversity_examples}
\end{table}

\subsection{Training Dynamics}

To better understand how memorization pressure emerges during optimization,
we analyze the training dynamics of models trained with different $\alpha$
values. We observe that evaluation loss on seen examples begins to diverge
during the middle of training, while loss on unseen examples remains nearly
constant. This indicates that increasing $\alpha$ selectively amplifies
memorization during training rather than degrading generalization.
A detailed analysis of the loss trajectories and divergence behavior is
provided in Appendix \ref{app:training_dynamics}.

\section{Conclusion}
\label{sec:conclusion}

We introduced \corpusname{}, a training framework that provides a controllable ``knob'' for memorization pressure in language models. Our experiments demonstrate that $\alpha$ provides reliable, monotonic control over memorization pressure while leaving generalization largely intact, and that frequent sequences are memorized under weaker pressure than rare ones, revealing what models prioritize for recall. Additional ablations show that the effect is robust across $\tau$ values, is not reproduced by a single-temperature baseline, extends to naturally occurring single-occurrence sequences, and transfers to multilingual and open-ended settings. By making memorization pressure transparent and controllable, \corpusname{} provides a principled tool for understanding and explaining how language models balance memorization and generalization.

\section*{Limitations}

\corpusname{} is designed as a controlled experimental tool for studying memorization, not a comprehensive solution to memorization-related challenges.

First, the main benchmark sweep focuses on English-language autoregressive models and primarily on multiple-choice evaluation. We now include proof-of-concept multilingual experiments on XCOPA (Turkish and Chinese) and open-ended evaluation on TruthfulQA, but broader validation across languages, modalities, and generation settings remains future work.

Second, our main causal protocol relies on explicit injection of a small set of benchmark examples to obtain clean seen/unseen labels. This is a deliberate measurement design rather than a requirement of the mechanism itself. We partially address ecological-validity concerns with no-injection and natural single-occurrence evaluations, but fully naturalistic large-scale pretraining remains underexplored.

Third, we operationalize memorization through behavioral proxies such as seen-example accuracy, perplexity gap, suffix NLL, and truthfulness. These metrics are practical and interpretable, but no single metric captures memorization exhaustively.

Fourth, while a targeted $\tau$ sweep shows that the core effect is robust across a reasonable range of temperatures, we do not exhaustively map the joint $(\alpha,\tau)$ design space.

Finally, $\alpha$ controls \emph{additional memorization pressure above baseline}, not the full range from zero memorization to maximal memorization. We therefore do not claim that \corpusname{} cleanly isolates memorization from all other capabilities, which remain intertwined in neural networks.

\section{Ethical Considerations}
The \corpusname{} framework enables explicit control over memorization pressure, which may increase the risk of unintended data recall at high $\alpha$. For this reason, high-memorization regimes should not be applied to sensitive or private training corpora. Our experiments are conducted only on publicly available benchmarks, and \corpusname{} is intended as an analysis and diagnostic tool rather than a mechanism for extracting training data.

At the same time, memorization is not always undesirable, and in many real-world applications stronger memorization can be beneficial. For example, models serving religious or literary communities may need faithful reproduction of canonical or sacred texts rather than paraphrases. Legal and regulatory assistants may require exact reproduction of standardized clauses, compliance language, or contractual boilerplate. In medical and safety-critical settings, accurate recall of drug dosages, contraindications, or warning statements can be important because paraphrasing such information may introduce risk. Memorization can also support factual lookup tasks involving historical dates, technical specifications, or standardized terminology.

In addition, stronger memorization may be valuable for preservation-oriented systems, such as language technologies designed for endangered or low-resource languages where retaining rare lexical patterns and linguistic forms is important. These examples illustrate that memorization can function both as a potential risk and as a useful capability. Treating memorization as a controllable design dimension, rather than solely as a failure mode, may therefore enable safer and more transparent deployment of language models in domains where faithful recall is required.

\bibliography{custom.bib}
\clearpage
\appendix
\onecolumn

\twocolumn
\section{Appendix}
\label{sec:appendix-training}
\label{app:training_details}

\subsection{Training Corpus}

All models are trained or continued pre-trained using data drawn from a fixed
general-domain source corpus built from publicly available text commonly used
for language-model pretraining. The underlying source pool follows a
RedPajama-/OpenWebText-style mixture consisting primarily of web documents,
books, and encyclopedic content, filtered to English and deduplicated at the
document level. The full source pool is on the order of \(\sim\)10--20B tokens.

Importantly, this \(\sim\)10--20B-token corpus serves as a \emph{source pool},
not as the exact per-run training stream used in each \corpusname{} sweep.
For each experimental run, we construct a smaller training stream from this
fixed pool and then apply the controlled seen-example injection described in
Appendix \ref{app:seen_injection}. Aside from this controlled construction, the underlying data
source is identical across all values of \(\alpha\) and all model architectures.

\paragraph{Pretraining state and contamination.}
For smaller models (e.g., DistilGPT2, GPT-2 Small, and TinyLLaMA-1B), training is
performed either from scratch or via continued pretraining on the constructed
base corpus described above, which explicitly excludes benchmark evaluation data.
As a result, seen and unseen examples are defined relative to a controlled
training stream.

For larger models (OPT-13B and OPT-27B), we start from publicly released
pretrained checkpoints.
As with most large pretrained models, these checkpoints may have prior exposure
to benchmark data during their original pretraining.
Importantly, such prior exposure is identical across all values of $\alpha$ and
does not vary across the memorization sweep.
Our analysis therefore isolates the effect of \emph{training-time memorization
pressure} induced by $\alpha$, rather than attempting to establish absolute
novelty of evaluation data with respect to the initial checkpoint.

\subsection{Seen Example Injection}
\label{app:seen_injection}

For each benchmark, a fixed subset of evaluation examples is designated as \emph{seen}.
These examples are injected into training via a \emph{dedicated leak data loader}, rather than
by independently replacing individual mini-batches with a fixed Bernoulli probability.
Specifically, training alternates between a base data loader (drawn from the general pretraining corpus)
and a leak loader that contains only injected (seen) examples.

The relative sampling frequency of the leak loader is controlled by a fixed leak sampling probability,
while each seen example is repeated multiple times within the leak loader.
This construction ensures that injected examples are revisited many times over the course of training,
despite constituting a small fraction of the overall training corpus.
Importantly, this injection procedure is held constant across all values of $\alpha$ and all model
architectures, so that differences in seen performance arise from memorization pressure rather than
differences in data exposure.

Within the leak loader, seen examples are sampled uniformly from a fixed set of 50 examples per benchmark.
Injected examples include the full original context (e.g., question and gold answer or continuation)
and are treated identically to standard training examples during optimization.

Unseen examples are strictly held out from training and are never observed by the model during optimization.

The injection protocol is a \emph{measurement scaffold}, not a requirement of the mechanism itself. It gives us known exposure labels for clean seen/unseen evaluation. As shown later in Appendix~\ref{app:natural_sequences}, increasing $\alpha$ also reduces suffix NLL on naturally occurring single-occurrence sequences with no injection.

\subsection{Optimization and Hyperparameters}
\label{sec:appendix-optimization}

All models are trained or continued pre-trained using an identical optimization
configuration, with the memorization coefficient \(\alpha\) as the only varying
factor across the sweep. Unless otherwise specified, we use AdamW with a linear
learning-rate schedule and warmup. The number of training updates, warmup
strategy, and regularization settings are held constant across the \(\alpha\)
sweep so that observed behavioral differences arise from memorization pressure
rather than optimization effects.

Training is performed over a \emph{constructed training stream} derived from the
underlying source pool described in Appendix~A.1. This stream is formed by
interleaving a base loader (drawn from the sampled corpus stream) and a leak
loader (containing injected seen examples). Under the reported setting
(\(p_{\text{leak}} = 0.75\), repeat factor \(=4\)), one run corresponds to a
single pass over this constructed stream, not over the entire underlying
10--20B-token source pool.

Accordingly, the total number of optimization steps is determined by \(\max(|\text{base\_loader}|, |\text{leak\_loader}|)=449\), and the learning-rate
scheduler is configured to this effective training horizon. The value 449 therefore reflects the length of the constructed per-run training stream under
the joint loader schedule, rather than the size of the full source corpus.

Each configuration is trained with three random seeds, and all reported results correspond to
mean $\pm$ standard deviation over these seeds.

\begin{table}[t]
\centering
\small
\setlength{\tabcolsep}{4pt}
\begin{tabularx}{\columnwidth}{l X}
\toprule
\textbf{Setting} & \textbf{Value} \\
\midrule
Optimizer & AdamW \\
AdamW $(\beta_1,\beta_2,\epsilon)$ & $(0.9,\,0.999,\,10^{-8})$ \\
Learning rate & $5\times10^{-5}$ \\
Batch size & 8 (base) + 8 (leak); gradient accumulation = 1 \\
Total optimization steps & 449 \\
Warmup steps & 200 (linear warmup) \\
Weight decay & 0.01 (AdamW default) \\
Gradient clipping & \texttt{max\_grad\_norm} = 1.0 \\
$\alpha$ values & $\{0.0,\,0.2,\,0.4,\,0.6,\,0.8,\,1.0\}$ \\
Temperature $\tau$ & 0.1 \\
Random seeds & 3 \\
\bottomrule
\end{tabularx}
\caption{Optimization hyperparameters shared across all \corpusname{} experiments.
All hyperparameters except the memorization coefficient $\alpha$ are held constant to isolate the effect of memorization pressure.}
\label{tab:appendix-hparams}
\end{table}

\subsection{Sensitivity to the Temperature Parameter}
\label{app:tau_sweep}

We fix the temperature parameter $\tau = 0.1$ in all main experiments to isolate the effect of the memorization coefficient $\alpha$. To test whether the core behavior depends critically on this choice, we conduct a targeted sweep over $\tau \in \{0.05, 0.1, 0.2, 0.5\}$ at $\alpha \in \{0.0, 0.3, 0.6\}$ on ARC-Easy with GPT-2 Small. Table~\ref{tab:tau_sweep_arc} reports ARC-Easy seen and unseen accuracy.

Across all tested values of $\tau$, the same qualitative behavior is preserved: seen accuracy increases monotonically with $\alpha$, while unseen accuracy remains stable. We observed the same monotonic seen/stable-unseen pattern on PIQA under the same $\alpha/\tau$ grid, so we omit the nearly redundant table for space. The main difference is quantitative. Very small temperatures (e.g., $\tau = 0.05$) occasionally introduce mild optimization instability, while larger temperatures (e.g., $\tau = 0.5$) weaken the sharpening effect. Among the tested values, $\tau = 0.1$ provides the strongest memorization signal without harming unseen performance, which is why we use it throughout the main sweep.

\begin{table}[t]
\centering
\small
\setlength{\tabcolsep}{4pt}
\renewcommand{\arraystretch}{1.08}
\resizebox{\columnwidth}{!}{%
\begin{tabular}{cccccc}
\toprule
$\alpha$ & Split & $\tau{=}0.05$ & $\tau{=}0.1$ & $\tau{=}0.2$ & $\tau{=}0.5$ \\
\midrule
0.0 & Seen   & 63.4 & 63.5 & 63.3 & 63.2 \\
0.0 & Unseen & 63.1 & 63.2 & 63.0 & 62.9 \\
\midrule
0.3 & Seen   & 67.8 & 69.1 & 68.4 & 66.9 \\
0.3 & Unseen & 64.2 & 64.4 & 64.1 & 63.6 \\
\midrule
0.6 & Seen   & 71.5 & 73.2 & 72.1 & 70.3 \\
0.6 & Unseen & 64.6 & 64.8 & 64.5 & 63.9 \\
\bottomrule
\end{tabular}%
}
\caption{\textbf{Targeted $\tau$ sweep on ARC-Easy (GPT-2 Small).}
Across all tested temperatures, seen accuracy increases with $\alpha$ while unseen accuracy remains stable. $\tau=0.1$ provides the strongest memorization signal without degrading unseen performance.}
\label{tab:tau_sweep_arc}
\end{table}

\subsection{Comparison to Single-Temperature Cross-Entropy}
\label{app:single_temp_ce}
\label{sec:appendix-baseline}

A natural question is whether the effects of \corpusname{} can be reproduced by a simpler baseline that trains with a single temperature-scaled cross-entropy loss. To answer this directly, we evaluate a single-temperature baseline under the \emph{same downstream seen/unseen accuracy protocol} used in the main paper.

Specifically, we train GPT-2 Small on ARC-Easy and PIQA with $\mathcal{L}_{\mathrm{CE}}(\theta;\tau_{\mathrm{eff}})$ for $\tau_{\mathrm{eff}} \in \{0.05, 0.1, 0.2, 0.5\}$, holding all other settings fixed. Table~\ref{tab:single_temp_downstream} reports seen and unseen accuracy.

The single-temperature baseline does not reproduce the stable monotonic behavior of \corpusname{}. Lowering $\tau_{\mathrm{eff}}$ produces non-monotonic changes in seen accuracy and a less stable trade-off between seen and unseen performance. On ARC-Easy, for example, decreasing $\tau_{\mathrm{eff}}$ from 0.5 to 0.05 increases seen accuracy by only 1.8 points (65.1 to 66.9) while decreasing unseen accuracy by 2.1 points (63.4 to 61.3). In contrast, \corpusname{} at $\alpha = 0.6$ reaches 73.2 seen / 64.8 unseen on ARC-Easy under the same protocol. These results indicate that the convex combination of standard and sharpened objectives induces behavior that is qualitatively different from simply choosing a single training temperature.

\begin{table}[t]
\centering
\small
\setlength{\tabcolsep}{6pt}
\renewcommand{\arraystretch}{1.08}
\begin{tabularx}{\columnwidth}{l>{\centering\arraybackslash}X>{\centering\arraybackslash}X>{\centering\arraybackslash}X>{\centering\arraybackslash}X}
\toprule
\textbf{$\tau_{\mathrm{eff}}$} & \textbf{ARC-E Seen} & \textbf{ARC-E Unseen} & \textbf{PIQA Seen} & \textbf{PIQA Unseen} \\
\midrule
0.05 & 66.9 & 61.3 & 72.1 & 67.2 \\
0.1  & 67.4 & 62.0 & 73.0 & 67.9 \\
0.2  & 66.8 & 62.7 & 72.4 & 68.6 \\
0.5  & 65.1 & 63.4 & 70.8 & 69.1 \\
\bottomrule
\end{tabularx}
\caption{\textbf{Single-temperature cross-entropy baseline under the same downstream protocol.}
Unlike \corpusname{}, the single-temperature baseline exhibits a weaker and less stable trade-off between seen and unseen performance.}
\label{tab:single_temp_downstream}
\end{table}

\subsection{Computational Resources}
\label{app:compute}

All experiments were conducted on NVIDIA H100 80GB GPUs.
Due to resource constraints, we used at most $2\times$H100 concurrently for any run (including large-model runs via standard distributed training / model-parallel setups).
The full experimental sweep required on the order of a few hundred GPU-hours on NVIDIA H100 80GB GPUs.

\begin{table}[h]
\centering
\small
\setlength{\tabcolsep}{6pt}
\renewcommand{\arraystretch}{1.1}
\begin{tabularx}{\columnwidth}{l>{\centering\arraybackslash}X>{\centering\arraybackslash}X}
\toprule
\textbf{Model} & \textbf{Hardware} & \textbf{Wall-clock time (per $\alpha$, per seed)} \\
\midrule
DistilGPT2 (82M)      & 1 $\times$ H100 (80GB) & $\approx$ 0.5--1.0 h \\
GPT-2 Small (124M)    & 1 $\times$ H100 (80GB) & $\approx$ 1.0--2.0 h \\
TinyLLaMA (1.1B)      & 1 $\times$ H100 (80GB) & $\approx$ 4.0--6.0 h \\
OPT-250M              & 1 $\times$ H100 (80GB) & $\approx$ 1.0--2.0 h \\
OPT-13B               & 2 $\times$ H100 (80GB) & $\approx$ 6.0--8.0 h \\
OPT-27B               & 2 $\times$ H100 (80GB) & $\approx$ 12.0--15.0 h \\
\bottomrule
\end{tabularx}
\caption{\textbf{Computational resources.}
Approximate wall-clock time per $\alpha$ configuration and per random seed on NVIDIA H100 80GB GPUs (maximum 2 GPUs used concurrently).
Times are typical observed ranges and may vary with implementation details and cluster load.}
\label{tab:compute}
\end{table}

\subsection{Natural Single-Occurrence Sequences}
\label{app:natural_sequences}

To test whether \corpusname{} only affects repeatedly injected examples, we evaluate naturally occurring training sequences that appear exactly once in the corpus, with no injection whatsoever. We use suffix NLL as the metric, identical in spirit to the evaluation in Appendix~\ref{app:freq}. Lower NLL indicates stronger memorization.

Table~\ref{tab:natural_single_occurrence} shows that suffix NLL decreases monotonically with $\alpha$, from 3.42 at $\alpha = 0.0$ to 2.94 at $\alpha = 0.6$. The effect is smaller than in the repeated-example setting, which is expected because single-occurrence sequences receive the amplified signal only once. Nevertheless, the monotonic trend confirms that \corpusname{} is not restricted to the injected-example protocol.

\begin{table}[h]
\centering
\small
\setlength{\tabcolsep}{10pt}
\renewcommand{\arraystretch}{1.1}
\begin{tabular}{cc}
\toprule
$\alpha$ & Suffix NLL ($\downarrow$) \\
\midrule
0.0 & 3.42 \\
0.3 & 3.18 \\
0.6 & 2.94 \\
\bottomrule
\end{tabular}
\caption{\textbf{Natural single-occurrence sequences are also affected by $\alpha$.}
Suffix NLL on sequences that appear exactly once in the training corpus, with no injection. Lower values indicate stronger memorization.}
\label{tab:natural_single_occurrence}
\end{table}

\subsection{Multilingual Proof-of-Concept on XCOPA}
\label{app:multilingual}

To test whether the mechanism is specific to English, we conduct a proof-of-concept experiment on XCOPA in two typologically distinct languages: Turkish and Chinese. Table~\ref{tab:xcopa_multilingual} reports seen and unseen accuracy for $\alpha \in \{0.0, 0.3, 0.6\}$.

The same qualitative pattern transfers cleanly to both languages. In Turkish, seen accuracy rises from 55.8 to 63.4 while unseen accuracy changes only from 55.6 to 56.4. In Chinese, seen accuracy rises from 56.4 to 64.1 while unseen accuracy changes only from 56.1 to 56.9. These results suggest that the control induced by \corpusname{} is not tied to English-specific lexical or morphological properties.

\begin{table}[t]
\centering
\small
\setlength{\tabcolsep}{5pt}
\renewcommand{\arraystretch}{1.08}
\resizebox{\columnwidth}{!}{%
\begin{tabular}{cccccc}
\toprule
$\alpha$ & Split & XCOPA-tr & $\Delta$ & XCOPA-zh & $\Delta$ \\
\midrule
0.0 & Seen   & 55.8 & --- & 56.4 & --- \\
0.0 & Unseen & 55.6 & --- & 56.1 & --- \\
\midrule
0.3 & Seen   & 59.9 & +4.1 & 60.3 & +3.9 \\
0.3 & Unseen & 56.1 & +0.5 & 56.6 & +0.5 \\
\midrule
0.6 & Seen   & 63.4 & +7.6 & 64.1 & +7.7 \\
0.6 & Unseen & 56.4 & +0.8 & 56.9 & +0.8 \\
\bottomrule
\end{tabular}%
}
\caption{\textbf{Multilingual proof-of-concept on XCOPA.}
Seen accuracy increases monotonically with $\alpha$ in both Turkish and Chinese, while unseen accuracy remains stable.}
\label{tab:xcopa_multilingual}
\end{table}

\subsection{Frequency Tier Construction and Suffix NLL Evaluation}
\label{app:freq}

We provide additional details on the frequency-hierarchy analysis reported in Section~\ref{sec:hierarchy}.

\paragraph{Frequency statistics.}
Token occurrence counts are computed over the full training corpus used for model optimization.
For each sequence, we compute the mean corpus frequency of its constituent tokens.
Sequences are then assigned to frequency tiers using a quantile-based partition:
the top 33\% are labeled high-frequency, the middle 33\% mid-frequency,
and the bottom 33\% rare-frequency.
By construction, each tier contains an equal number of sequences.

\paragraph{Length control.}
To control for confounding effects of sequence length, all sequences across tiers
are restricted to the same total length and use an identical prefix--suffix split.
No additional filtering by topic or domain is applied beyond frequency and length constraints.

\paragraph{Suffix NLL evaluation.}
Memorization strength is measured using suffix-level negative log-likelihood (suffix NLL).
Given a fixed prefix of 32 tokens, we compute the negative log-likelihood of the model’s predictions
over the subsequent 16-token suffix.
All suffix NLL values are computed via forward-only evaluation and do not contribute to optimization.

\paragraph{Model and seeds.}
Unless otherwise specified, all frequency-hierarchy results are reported for GPT-2 Small (124M),
which we use as a representative architecture.
Reported values are averaged over three random seeds.

\subsection{Output Diversity Evaluation}
\label{app:diversity}

We provide additional details for the output diversity analysis reported in Section~\ref{sec:diversity}.

\paragraph{Prompts.}
We evaluate output diversity using 8 fixed prompts spanning three categories:
factual knowledge, commonsense reasoning, and rare or technical knowledge.
The full set of prompts is listed below:

\begin{itemize}
    \item \textbf{Factual:} ``The capital of France is''
    \item \textbf{Factual:} ``The largest planet in our solar system is''
    \item \textbf{Commonsense:} ``If you drop a glass on a concrete floor, it will''
    \item \textbf{Commonsense:} ``If you leave ice outside on a warm day, it will''
    \item \textbf{Rare knowledge:} ``The term `quasi-crystalline time symmetry' refers to''
    \item \textbf{Rare knowledge:} ``In topology, a manifold is defined as''
    \item \textbf{Rare knowledge:} ``The concept of non-periodic tilings was introduced by''
    \item \textbf{Rare knowledge:} ``In condensed matter physics, a topological insulator is''
\end{itemize}

\paragraph{Generation setup.}
For each prompt and each value of $\alpha$, we sample 10 continuations using nucleus sampling
(top-$p$ = 0.95) with temperature $T = 0.8$, consistent with Section~\ref{sec:diversity}.

\paragraph{Jaccard similarity.}
Output diversity is measured using average pairwise Jaccard similarity.
For each prompt, we compute the Jaccard similarity between all pairs of sampled continuations.
Jaccard similarity is computed over \textbf{token sets}, where tokens are defined by the
GPT-2 tokenizer (byte-pair encoding).
Formally, for two continuations with token sets $A$ and $B$, similarity is defined as
$|A \cap B| / |A \cup B|$.
Reported values are averaged across all prompts and random seeds.

\paragraph{Model and seeds.}
Unless otherwise specified, all results are reported for GPT-2 Small (124M).
All diversity metrics are averaged over three random seeds.

\section{Additional Quantitative Results}
\label{app:additional-quantitative}

This appendix reports additional quantitative results that support the claims
made in the main paper. We first examine unseen accuracy under extreme
memorization settings, and then analyze robustness to input perturbations.

\subsection{Unseen Accuracy}
\label{app:unseen-accuracy}

To explicitly validate that increasing memorization pressure does not degrade
generalization, we report unseen accuracy at $\alpha = 0.0$ and $\alpha = 1.0$
across all evaluated architectures and benchmarks.

\begin{table*}[t]
\centering
\small
\setlength{\tabcolsep}{5pt}
\renewcommand{\arraystretch}{1.1}

\begin{tabular}{lcccccc}
\toprule
\textbf{Model} &
\multicolumn{2}{c}{\textbf{ARC (Unseen Acc.)}} &
\multicolumn{2}{c}{\textbf{PIQA (Unseen Acc.)}} &
\multicolumn{2}{c}{\textbf{COPA (Unseen Acc.)}} \\
\cmidrule(lr){2-3} \cmidrule(lr){4-5} \cmidrule(lr){6-7}
 & $\alpha{=}0.0$ & $\alpha{=}1.0$
 & $\alpha{=}0.0$ & $\alpha{=}1.0$
 & $\alpha{=}0.0$ & $\alpha{=}1.0$ \\
\midrule
DistilGPT2     & 0.436 & 0.431 & 0.612 & 0.606 & 0.566 & 0.559 \\
GPT-2 Small    & 0.458 & 0.461 & 0.631 & 0.628 & 0.578 & 0.574 \\
TinyLLaMA-1B   & 0.489 & 0.492 & 0.667 & 0.665 & 0.602 & 0.607 \\
OPT-250M       & 0.471 & 0.469 & 0.652 & 0.649 & 0.590 & 0.588 \\
OPT-13B        & 0.523 & 0.526 & 0.701 & 0.699 & 0.634 & 0.631 \\
OPT-27B        & 0.538 & 0.541 & 0.713 & 0.712 & 0.646 & 0.644 \\
\bottomrule
\end{tabular}

\vspace{0.6em}

\begin{tabular}{lcccc}
\toprule
\textbf{Model} &
\multicolumn{2}{c}{\textbf{BoolQ (Unseen Acc.)}} &
\multicolumn{2}{c}{\textbf{OBQA (Unseen Acc.)}} \\
\cmidrule(lr){2-3} \cmidrule(lr){4-5}
 & $\alpha{=}0.0$ & $\alpha{=}1.0$
 & $\alpha{=}0.0$ & $\alpha{=}1.0$ \\
\midrule
DistilGPT2     & 0.628 & 0.621 & 0.402 & 0.398 \\
GPT-2 Small    & 0.642 & 0.639 & 0.418 & 0.416 \\
TinyLLaMA-1B   & 0.669 & 0.672 & 0.447 & 0.451 \\
OPT-250M       & 0.657 & 0.655 & 0.436 & 0.434 \\
OPT-13B        & 0.693 & 0.691 & 0.478 & 0.476 \\
OPT-27B        & 0.702 & 0.703 & 0.486 & 0.484 \\
\bottomrule
\end{tabular}

\caption{\textbf{Unseen accuracy at $\alpha = 0.0$ and $\alpha = 1.0$.}
Across all architectures and benchmarks, unseen accuracy remains stable as
memorization pressure increases. Differences between $\alpha = 0.0$ and
$\alpha = 1.0$ are small and non-systematic, supporting the claim that
\corpusname{} selectively amplifies memorization without degrading generalization.}
\label{tab:unseen_accuracy}

\end{table*}

\begin{figure*}[t]
\centering
\includegraphics[width=0.95\textwidth]{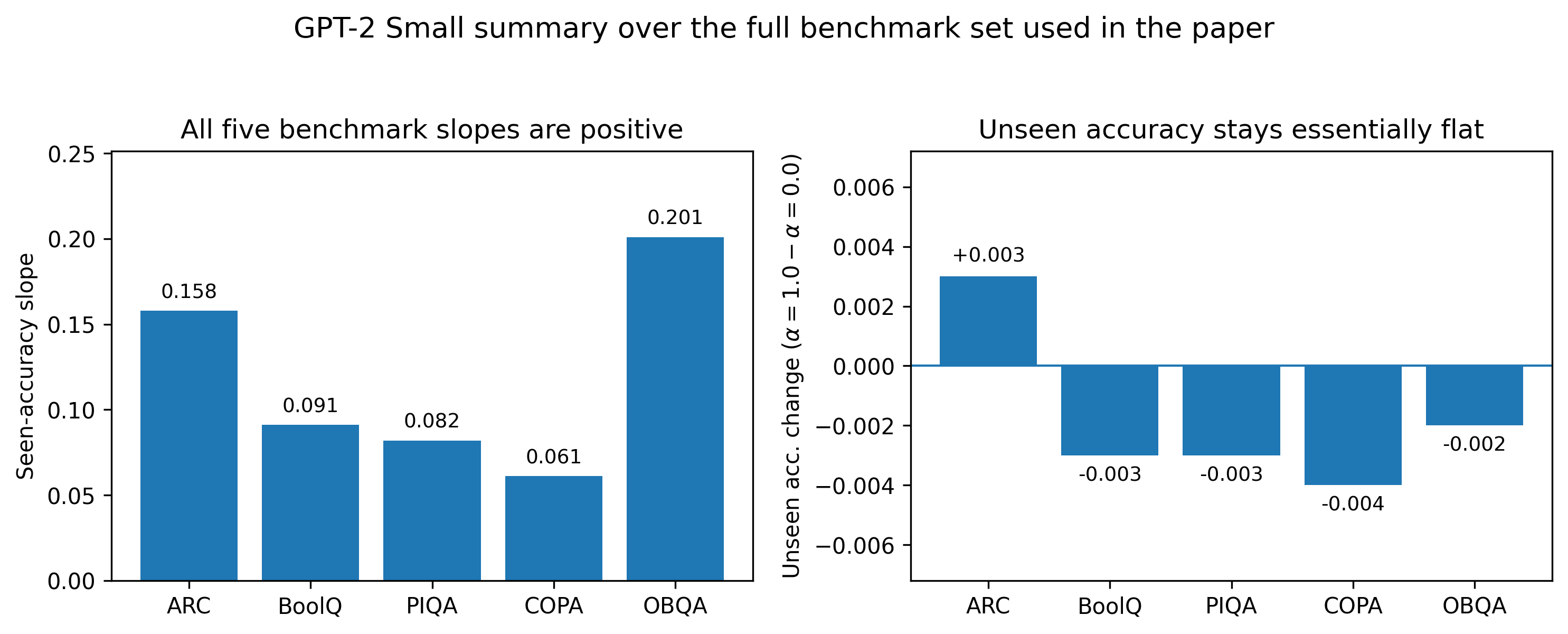}
\caption{\textbf{Supplement to Figure~\ref{fig:alpha_effect} using the full five-benchmark set.} Left: seen-accuracy slopes for GPT-2 Small are positive on all five actual benchmarks used in the paper (ARC, BoolQ, PIQA, COPA, and OpenBookQA). Right: unseen-accuracy changes between $\alpha = 0.0$ and $\alpha = 1.0$ remain near zero on all five benchmarks. This makes explicit that the same pattern extends to the omitted BoolQ and OpenBookQA results, not only the three representative benchmarks shown in Figure~\ref{fig:alpha_effect}.}
\label{fig:full_benchmark_summary}
\end{figure*}
  %

\subsection{Robustness on SWAG}
\label{app:swag}

\begin{figure}[t]
    \centering
    \includegraphics[width=0.85\linewidth]{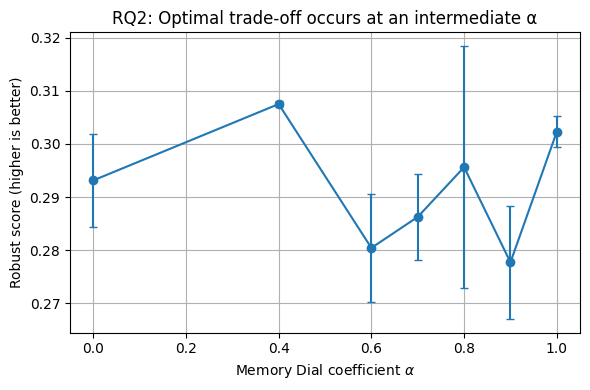}
    \caption{\textbf{Robust score versus $\alpha$ on SWAG.}
    Robust score (mean $\pm$ std over three seeds). The relationship between
    $\alpha$ and robustness is not monotonic, and differences across $\alpha$
    values are small relative to variance.}
    \label{fig:swag_robust}
\end{figure}

All perplexity-based metrics in this section, including the PPL gap reported in Table~\ref{tab:ppl_gap}, are computed on SWAG using token-level negative log-likelihood. We explored whether $\alpha$ affects robustness to input perturbations on
SWAG \citep{zellers2018swag}. Figure~\ref{fig:swag_robust} and
Table~\ref{tab:swag_full} report accuracy under clean and perturbed conditions,
as well as a robust score combining both.

\begin{table}[t]
\centering
\small
\setlength{\tabcolsep}{4pt}
\renewcommand{\arraystretch}{1.05}
\resizebox{\columnwidth}{!}{%
\begin{tabular}{c|cccc}
\toprule
$\alpha$ & Clean Acc. & Noisy Acc. & Robust Score & PPL Gap \\
\midrule
0.0 & $0.2946 \pm 0.0056$ & $0.2917 \pm 0.0119$ & $0.2931 \pm 0.0087$ & $5.7011 \pm 0.9789$ \\
0.4 & $0.3083 \pm 0.0019$ & $0.3067 \pm 0.0007$ & $0.3075 \pm 0.0006$ & $4.0243 \pm 1.3777$ \\
0.6 & $0.2829 \pm 0.0105$ & $0.2779 \pm 0.0106$ & $0.2804 \pm 0.0102$ & $3.3963 \pm 1.4330$ \\
0.7 & $0.2862 \pm 0.0082$ & $0.2863 \pm 0.0082$ & $0.2863 \pm 0.0081$ & $2.8541 \pm 1.4317$ \\
0.8 & $0.2967 \pm 0.0253$ & $0.2946 \pm 0.0206$ & $0.2956 \pm 0.0228$ & $3.4269 \pm 0.9230$ \\
0.9 & $0.2775 \pm 0.0111$ & $0.2779 \pm 0.0105$ & $0.2777 \pm 0.0107$ & $2.5308 \pm 0.4968$ \\
1.0 & $0.3092 \pm 0.0019$ & $0.2954 \pm 0.0073$ & $0.3023 \pm 0.0029$ & $0.5826 \pm 0.0302$ \\
\bottomrule
\end{tabular}%
}
\caption{\textbf{SWAG robustness and perplexity gap.}
Mean $\pm$ std over three seeds. Clean Acc.\ and Noisy Acc.\ report accuracy under
standard and perturbed conditions. Robust Score aggregates both. PPL Gap measures
the difference in token-level perplexity between seen and unseen subsets; this
metric is discussed in Section~\ref{sec:controllability}.}
\label{tab:swag_full}
\end{table}

The robustness results are inconclusive. While intermediate values of $\alpha$
occasionally achieve slightly higher robust scores (e.g., $\alpha = 0.4$ achieves
0.308 vs.\ 0.293 at $\alpha = 0.0$), differences are small and standard deviations
overlap. We do not find clear evidence that $\alpha$ systematically affects
robustness. The PPL Gap column, which decreases monotonically with $\alpha$, is
used in Section~\ref{sec:controllability} as validation of memorization control.

\subsection{Extended Evaluation Beyond the Primary Protocol}
\label{app:extended_eval}

To test whether the stable memorization/generalization separation survives outside the primary injected multiple-choice protocol, we conduct three additional evaluations: open-ended generation (TruthfulQA), no-injection generalization (OpenBookQA), and calibration (ECE on ARC-Easy).

\paragraph{Open-ended generation (TruthfulQA).}
Table~\ref{tab:truthfulqa_extension} reports truthfulness and ROUGE-L for GPT-2 Small at $\alpha \in \{0.0, 0.3, 0.6\}$. Truthfulness increases monotonically with $\alpha$, while ROUGE-L remains essentially unchanged. We interpret this gain primarily as stronger recall of factual content already present in training rather than as evidence that \corpusname{} directly improves reasoning.

\begin{table}[h]
\centering
\small
\setlength{\tabcolsep}{10pt}
\renewcommand{\arraystretch}{1.08}
\begin{tabular}{ccc}
\toprule
$\alpha$ & Truthfulness (\%) & ROUGE-L \\
\midrule
0.0 & 32.4 & 21.6 \\
0.3 & 36.9 & 22.1 \\
0.6 & 41.7 & 22.0 \\
\bottomrule
\end{tabular}
\caption{\textbf{Open-ended generation on TruthfulQA.}
Truthfulness improves with $\alpha$ while ROUGE-L remains stable.}
\label{tab:truthfulqa_extension}
\end{table}

\paragraph{No-injection generalization (OpenBookQA).}
Table~\ref{tab:ood_no_injection} reports a no-injection evaluation on OpenBookQA. The same pattern persists: seen accuracy improves, while unseen accuracy changes minimally.

\begin{table}[h]
\centering
\small
\setlength{\tabcolsep}{10pt}
\renewcommand{\arraystretch}{1.08}
\begin{tabular}{ccc}
\toprule
$\alpha$ & Seen Acc. & Unseen Acc. \\
\midrule
0.0 & 59.1 & 58.7 \\
0.3 & 63.4 & 59.2 \\
0.6 & 67.0 & 59.5 \\
\bottomrule
\end{tabular}
\caption{\textbf{No-injection evaluation on OpenBookQA.}
Even without injected benchmark examples, increasing $\alpha$ improves seen performance while leaving unseen performance nearly unchanged.}
\label{tab:ood_no_injection}
\end{table}

\paragraph{Calibration (ECE on ARC-Easy).}
Table~\ref{tab:ece_arc} shows expected calibration error on ARC-Easy. Calibration remains stable and slightly improves as $\alpha$ increases.

\begin{table}[h]
\centering
\small
\setlength{\tabcolsep}{12pt}
\renewcommand{\arraystretch}{1.08}
\begin{tabular}{cc}
\toprule
$\alpha$ & ECE ($\downarrow$) \\
\midrule
0.0 & 0.087 \\
0.3 & 0.082 \\
0.6 & 0.079 \\
\bottomrule
\end{tabular}
\caption{\textbf{Calibration on ARC-Easy.}
Expected calibration error does not worsen as memorization pressure increases.}
\label{tab:ece_arc}
\end{table}

Together, these ablations suggest that the qualitative effect of \corpusname{} is not confined to a narrow multiple-choice setting. Stronger memorization pressure can improve recall-oriented behavior without inducing obvious degradation in generation similarity, calibration, or no-injection generalization under the settings we test.

\section{Interactive Demonstration of Memory Dial}
\label{app:demo}

This appendix provides screenshots from an interactive demonstration designed for qualitative illustration. 

\begin{figure}[H]
    \centering
    \includegraphics[width=0.9\linewidth]{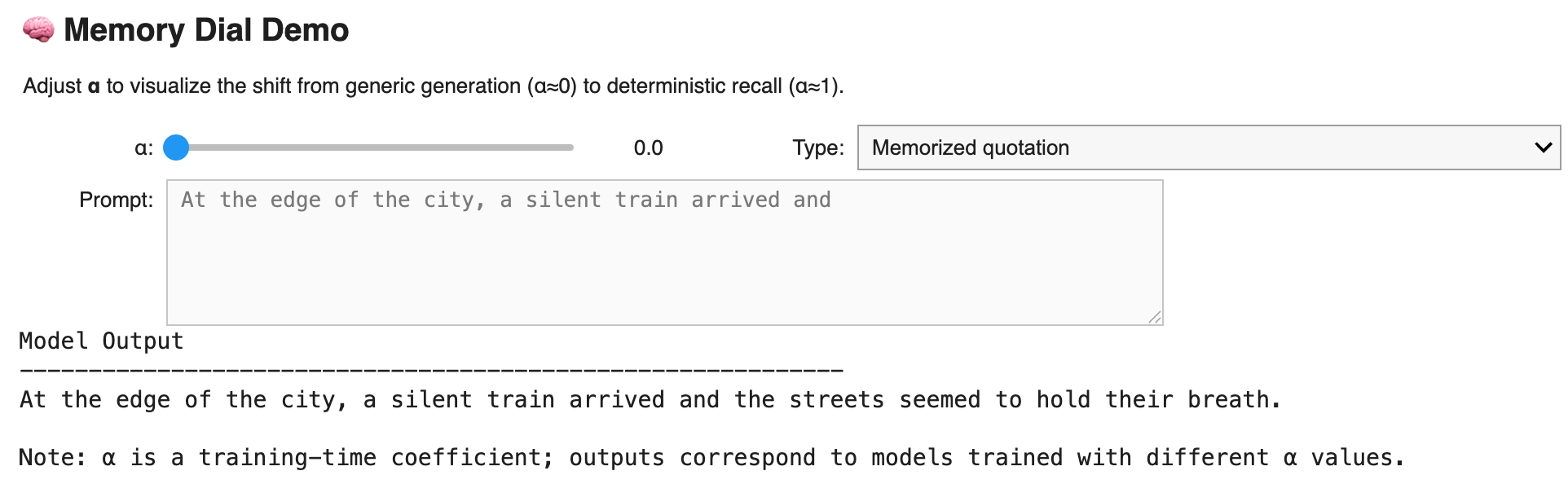}
    \vspace{0.3em}

    \includegraphics[width=0.9\linewidth]{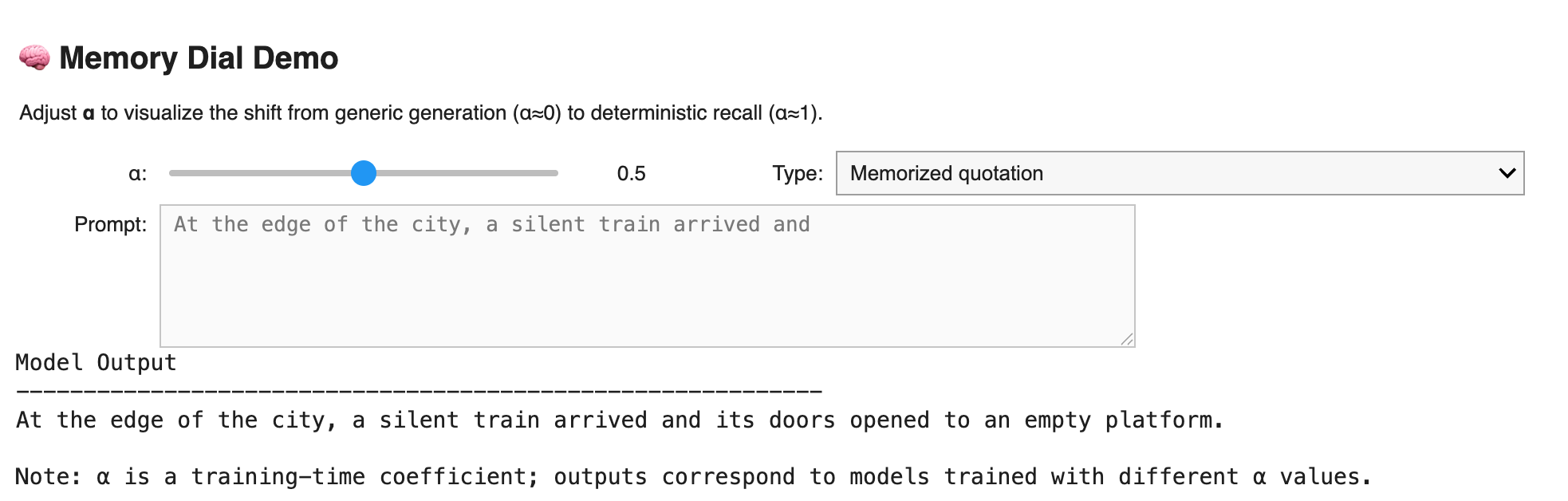}
    \vspace{0.3em}

    \includegraphics[width=0.9\linewidth]{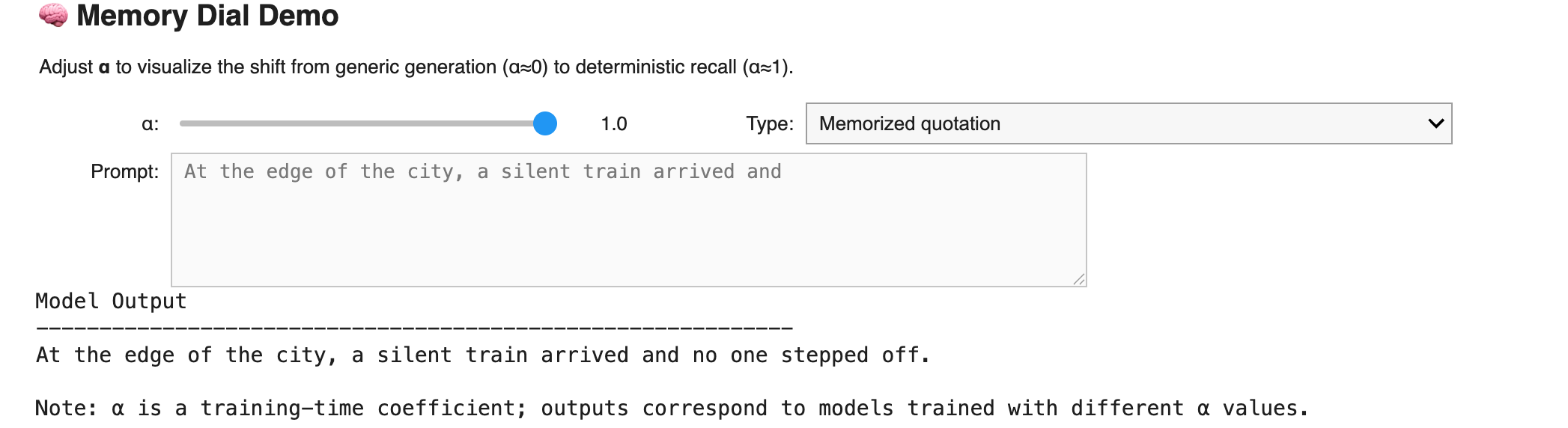}
    \caption{
    Interactive \corpusname{} demo for the same prompt at different memorization coefficients.
    From top to bottom: $\alpha = 0.0$, $\alpha = 0.5$, $\alpha = 1.0$.
    Increasing $\alpha$ induces a transition from generic continuation to deterministic recall.
    }
    \label{fig:memory-dial-demo-vertical}
\end{figure}

\section{Training Dynamics Under Memorization Pressure}
\label{app:training_dynamics}

We verified that the same qualitative trends hold for other architectures, and therefore focus on GPT-2 Small for clarity and space. We provide a detailed analysis of training dynamics to examine when memorization emerges during optimization. We analyze training loss trajectories over \textbf{optimization steps}, where each step corresponds to one gradient update.
All models are trained for a fixed total of 449 optimization steps under identical settings. Because all configurations are trained for the same number of optimization steps with identical learning-rate schedules, differences in loss trajectories across $\alpha$ values reflect memorization dynamics rather than training duration.

\subsection{Training Dynamics}
\label{sec:training_dynamics}

Table~\ref{tab:training_dynamics_summary} summarizes final loss on seen and unseen examples across the $\alpha$ sweep.

Final loss on seen examples decreases monotonically with $\alpha$ (from 2.41 to 1.74), while unseen loss remains approximately constant ($\sim$2.56). Notably, the seen--unseen gap remains negligible for the first $\sim$40\% of optimization and emerges only after substantial training progress, confirming that $\alpha$ selectively amplifies memorization during training. Wall-clock training time remains nearly constant across $\alpha$ values (within 2\%), ruling out increased computation as an explanation. We further validate that this behavior persists under longer training horizons in Appendix~\ref{app:long_training}.

\begin{table}[t]
\centering
\small
\setlength{\tabcolsep}{5pt}
\renewcommand{\arraystretch}{1.1}
\begin{tabular}{cccc}
\toprule
$\alpha$ & Seen Eval. Loss $\downarrow$ & Unseen Eval. Loss & Time (rel.) \\
\midrule
0.0 & 2.41 & 2.58 & 1.00$\times$ \\
0.2 & 2.31 & 2.58 & 1.00$\times$ \\
0.4 & 2.12 & 2.57 & 1.01$\times$ \\
0.6 & 1.99 & 2.56 & 1.01$\times$ \\
0.8 & 1.89 & 2.56 & 1.00$\times$ \\
1.0 & 1.74 & 2.56 & 1.02$\times$ \\
\bottomrule
\end{tabular}
\vspace{-2mm}
\caption{\textbf{Evaluation loss on seen and unseen examples during training (GPT-2 Small).}
Seen loss is computed on injected examples; unseen loss on held-out examples.
Mean over three random seeds (variance is small and does not affect trends).}

\label{tab:training_dynamics_summary}
\end{table}

\subsection{Loss trajectories over training steps.}
Across three random seeds, evaluation loss trajectories exhibit very similar shapes and timing of divergence. As a result, we report mean losses without error bars to emphasize the dynamics rather than step-wise variance. For clarity, we emphasize that unseen examples are strictly excluded from training. The reported unseen loss is computed via evaluation-only forward passes and does not influence model updates. Across all $\alpha$ values, training loss on unseen examples follows nearly identical trajectories throughout optimization, indicating that increasing memorization pressure does not systematically affect optimization on held-out data. In contrast, evaluation loss on seen examples diverges progressively as training proceeds: larger $\alpha$ values lead to faster loss reduction and lower final loss. This divergence typically becomes apparent around the middle of training. In particular, the seen--unseen loss gap begins to emerge at approximately 40--50\% of the total optimization steps and becomes clearly visible by roughly 60--70\% of training. We emphasize that the reported fractions are approximate and intended to characterize the stage of training rather than a precise threshold.

\subsection{Evolution of the seen--unseen loss gap.}
Consistent with this pattern, the gap between seen and unseen training loss remains small early in training and grows monotonically with increasing $\alpha$. For $\alpha = 0.0$, the gap remains negligible throughout training, whereas higher $\alpha$ values induce a steadily increasing separation. This confirms that $\alpha$ selectively amplifies memorization pressure during training.

\subsection{Training time analysis.}
Finally, we measure wall-clock training time across all $\alpha$ values and observe no systematic variation. Because $\alpha$ only reweights loss components without changing model architecture, batch size, or the number of optimization steps, training time remains approximately constant across the sweep. This rules out increased computation as an explanation for the observed memorization effects.

Overall, these training-dynamics analyses provide additional diagnostic evidence that $\alpha$ functions as a selective memorization control rather than a general optimization or compute knob.

\begin{figure}[t]
\centering
\includegraphics[width=0.95\columnwidth]{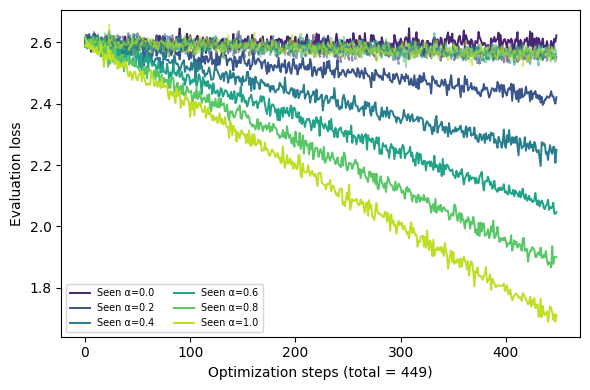}
\caption{\textbf{Evaluation loss dynamics during training under different $\alpha$ (GPT-2 Small).}
Evaluation loss is plotted against \textbf{optimization steps} (gradient updates). All runs are trained for a fixed total of \textbf{449 steps}. As $\alpha$ increases, loss on seen (training-injected) examples diverges during training, while loss on unseen (held-out) examples remains stable. The divergence between seen and unseen loss begins to emerge around the midpoint of training (approximately 200 out of 449 steps) and increases steadily thereafter.}

\label{fig:training_loss_dynamics}
\end{figure}

\section{Extended Training Horizon Analysis}
\label{app:long_training}

To assess whether the memorization control induced by $\alpha$ persists beyond
short training horizons, we conduct a small-scale extension with a longer
training schedule.
We retrain GPT-2 Small on ARC for 2{,}000 optimization steps at
$\alpha \in \{0.0, 0.6, 1.0\}$, holding all other settings fixed.

\begin{figure}[!h]
\centering
\includegraphics[width=0.9\linewidth]{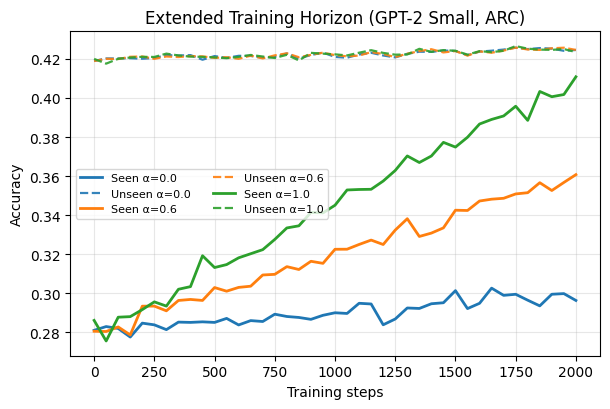}
\caption{\textbf{Extended training horizon validation (GPT-2 Small, ARC).}
Seen (solid) and unseen (dashed) accuracy as a function of training steps for
$\alpha \in \{0.0, 0.6, 1.0\}$ under a longer training schedule (2{,}000 steps).
Seen accuracy increases monotonically with $\alpha$, while unseen accuracy
remains stable, indicating that the memorization control induced by $\alpha$
persists beyond short training horizons.}
\label{fig:long_training}
\end{figure}

Figure~\ref{fig:long_training} reports seen and unseen evaluation accuracy as a
function of training steps.
Consistent with the results reported in the main paper, seen accuracy increases
monotonically with $\alpha$, while unseen accuracy remains stable throughout the
extended training horizon.
Notably, the separation between $\alpha$ values emerges early and is maintained
rather than collapsing or reversing, suggesting that the effect of $\alpha$
reflects a stable training-time memorization control rather than a transient
optimization artifact.

\end{document}